\documentclass{article}

\usepackage{arxiv}

\usepackage[utf8]{inputenc}
\usepackage[T1]{fontenc}
\usepackage{amsfonts}
\usepackage{nicefrac}
\usepackage{microtype}
\usepackage{natbib}

\usepackage{amsmath,amsfonts,bm}

\def\eqref#1{equation~\ref{#1}}

\def\1{\bm{1}}

\DeclareMathAlphabet{\mathsfit}{\encodingdefault}{\sfdefault}{m}{sl}
\SetMathAlphabet{\mathsfit}{bold}{\encodingdefault}{\sfdefault}{bx}{n}

\usepackage{hyperref}
\usepackage{url}
\usepackage{graphicx}
\usepackage{booktabs}
\usepackage{multirow}
\usepackage{placeins}
\usepackage{xcolor}
\usepackage{float}
\usepackage{wrapfig}

\usepackage[breakable]{tcolorbox}
\definecolor{neagent}{HTML}{B5442B}
\definecolor{neagentbg}{HTML}{F6E9E6}
\definecolor{neruntime}{HTML}{0B6E4F}
\definecolor{neruntimebg}{HTML}{E2EEEA}
\definecolor{nemeta}{HTML}{2E5C8A}
\definecolor{nemetabg}{HTML}{E6EBF1}
\tcbuselibrary{skins}
\newtcolorbox[auto counter, number within=section]{promptbox}[2][]{breakable,
  enhanced, frame hidden, colback=black!1,
  borderline west={1.4pt}{0pt}{black!45},
  colbacktitle=black!8, coltitle=black!80,
  left=8pt, right=7pt, top=5pt, bottom=5pt,
  toptitle=3pt, bottomtitle=3pt,
  fontupper=\small\sffamily, fonttitle=\small\bfseries\sffamily,
  title={Box~\thetcbcounter: #2}, #1}
\tcbset{
  searchprompt/.style={colback=neagentbg!24!white,
    borderline west={1.6pt}{0pt}{neagent}, coltitle=neagent!72!black,
    colbacktitle=neagentbg},
  runtimeprompt/.style={colback=neruntimebg!24!white,
    borderline west={1.6pt}{0pt}{neruntime}, coltitle=neruntime!72!black,
    colbacktitle=neruntimebg},
  metaprompt/.style={colback=nemetabg!28!white,
    borderline west={1.6pt}{0pt}{nemeta}, coltitle=nemeta!72!black,
    colbacktitle=nemetabg}
}

\newcommand{\faded}[1]{\textcolor{black!45}{#1}}

\usepackage[normalem]{ulem}

\title{AgentDiscover: Autonomous Discovery with \\ Minimal Search Scaffolding}

\author{Mahdi Farahbakhsh, Ilan Sela, Fatemeh Doudi, Vishnu Teja Kunde, \\
\textbf{Krishna Narayanan, Jean-Francois Chamberland, Dileep Kalathil} \\
Texas A\&M University \\
{\small\texttt{\{mahdi.farahbakhsh,isela1,fatemehdoudi,vishnukunde,krn,chmbrlnd,dileep.kalathil\}@tamu.edu}}
}

\date{}
\renewcommand{\undertitle}{}
\renewcommand{\headeright}{}
\renewcommand{\shorttitle}{AgentDiscover: Autonomous Discovery with Minimal Search Scaffolding}

\newcommand{\method}{AgentDiscover}

\begin{document}

\maketitle

\begin{figure}[h!]
\vspace{-1.5cm}
\centering
\includegraphics[width=0.93\linewidth]{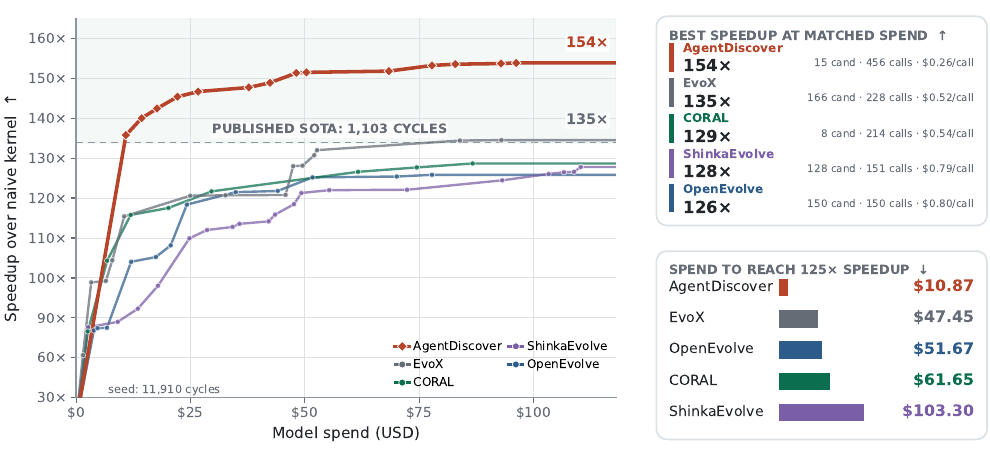}
\vspace{-0.5cm}
\caption{On the kernel builder task, \method{} reaches 960 cycles, the lowest
count among the frameworks we compare against, and it reaches comparable cycle
counts at lower model spend (more tasks in Appendix~\ref{app:spend-curves}).
All platforms run on Claude Opus~5 with the same budget.}
\label{fig:kernel}
\end{figure}

\begin{abstract}
Frameworks that use large language models for scientific discovery typically
rely on a fixed, human-designed algorithm that decides what the model sees at
each step, leaving the model only the role of proposer.
The model knows nothing of the search beyond what it is shown.
As models grow more capable, a question arises: does a search strategy chosen
by a human before the run scale better than promoting the model from proposer
to planner and letting it own the search?
\emph{The Bitter Lesson} suggests that choosing the strategy in advance is the
kind of hand-designed structure that general methods eventually outscale.
We introduce \method{}, in which a coding agent plans the search using its
context as working memory, runs experiments, and records every attempt in a
database of ideas, candidates, and their relations.
This database serves as the agent's long-term memory and is structured so
that the selection rules of classical algorithms such as MAP-Elites and Monte
Carlo tree search each reduce to a single query, which the agent is free to
use, combine, or replace.
A server maintains the database and steers the agent after every submission,
keeping it on course over long runs.
In our experiments, \method{} is more cost-efficient than existing frameworks,
reaching better scores at lower cost.
On tasks in kernel engineering, biology, algorithm design, and mathematics, AgentDiscover outperforms prior discovery frameworks. Its programs would have placed first among human competitors in seven past AtCoder heuristic contests, and on eleven mathematical and systems optimization tasks it matches or exceeds every baseline that uses the same model.
Our code is available \href{https://github.com/mhdfb/AgentDiscover}{here}.

\end{abstract}

\section{Introduction}
\label{sec:introduction}
\vspace{-0.5cm}
Large language models (LLMs) are increasingly capable problem solvers, but their solutions must be verified.
Returning each verdict to the model lets it improve its solutions, closing a loop of proposal and verification.
When both the candidate solution and its evaluator are programs, this loop needs no human intervention:
the model writes a program that either encodes a solution or searches for one, and an evaluator program runs it and scores the result.
This recipe, typically run as an evolutionary search, has already produced genuine mathematical and scientific discoveries.
For instance, FunSearch \citep{romera2024mathematical} found new constructions in extremal combinatorics, and AlphaEvolve \citep{novikov2025alphaevolve} improved the best known results on roughly a fifth of more than fifty open mathematical problems.
A growing family of open systems follows the same recipe \citep{openevolve,lange2026shinkaevolve,cemri2026adaevolve,liu2026evox}, and some go further by training the model on each problem at test time \citep{wang2025thetaevolve,yuksekgonul2026learning}.

These systems share a common architecture with three roles.
A \textit{search algorithm} selects past candidates for the model's context, a \textit{proposer}, a language model, writes a new candidate from them, and an \textit{evaluator} scores it; the result will then be recorded, typically in an archive of past candidates.
In most implementations, the selection rule is designed by humans and fixed before the run: AlphaEvolve combines MAP-Elites \citep{mouret2015illuminating} with island models, ShinkaEvolve \citep{lange2026shinkaevolve} extends a prescribed population loop with novelty filtering, and MCTS-AHD \citep{zheng2025mctsahd} organizes candidates in a tree and selects among them with the upper confidence bound for trees (UCT) \citep{kocsis2006bandit}.
More recent systems adapt this rule during a run, but only within a space prescribed beforehand by their designers \citep{cemri2026adaevolve,liu2026evox}.
In every case, the search is run by a human-designed algorithm and the LLM only proposes: it sees the candidates that the algorithm puts in its context, but has no say in which ones those are.

Leaving the search to a fixed human-designed algorithm becomes harder to justify as LLMs improve.
In \emph{The Bitter Lesson}, Sutton observes that methods built on human knowledge are eventually overtaken by general methods that exploit growing computation \citep{sutton2019bitter}.
Choosing the search strategy before the run is such a use of human knowledge; it bets that the designer understands the problem better than the language model will.
The LLM, meanwhile, could run the search itself: classical algorithms such as MAP-Elites and UCT are well documented in the text it is trained on, and \textit{coding agents}, which let an LLM write and run code with tools, can implement them when needed.
Knowing the alternatives matters because no single search strategy is best across problems \citep{wolpert1997nfl}.
In automated evolutionary discovery specifically, a budget-matched comparison of 30 fixed search strategies (``harnesses'') found none reliably better than a simple baseline across models and problems, and the full OpenEvolve recipe ranked last \citep{gupta2026harness}.
The best strategy can also change within a run \citep{cemri2026adaevolve,liu2026evox}; fixing one for the whole search leaves the model's knowledge of the alternatives unused.
This raises a natural question: \textit{can coding agents fill both roles, one planning the search and another proposing candidate solutions with minimal human-engineered search heuristics?}

Handing the search to a coding agent is not enough on its own.
The agent must remember what it has tried long after those attempts have left its context, retrieve the relevant part of that history cheaply, stay on course over hours of work, and do all of this without being able to tamper with the evaluator that scores it.
Without a structured memory, the agent must reconstruct the state of the search from raw logs before it can apply any strategy to it. Storing such a memory is cheap; reading it is not.
Every attempt can be recorded, with its score, its lineage, and the ideas it builds on, at negligible cost next to the compute a run already spends, but an agent pays in context for every record it reads. 
Databases have long addressed this: a query states what is wanted, and indexes and a query planner find it, so the reader receives only the answer. 
This raises a second question: \textit{can the history of a search be kept so that the agent retrieves only what it needs, at the cost of a query?}

\begin{figure}[t]
    \centering
    \includegraphics[width=\textwidth]{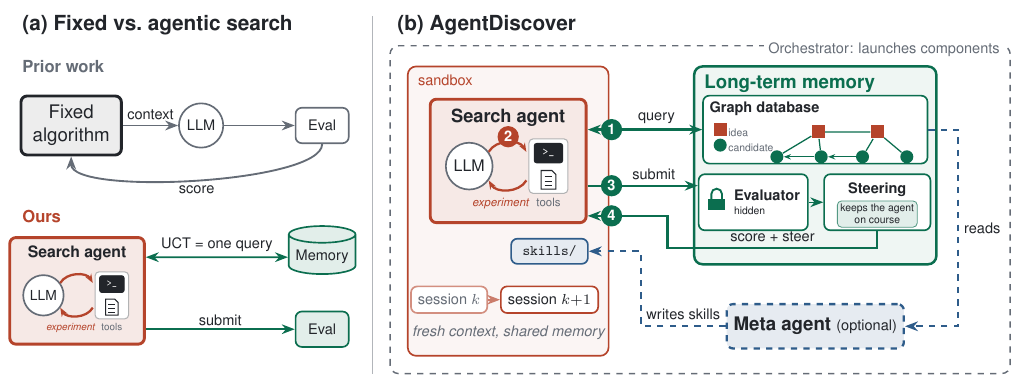}
    \vspace{-0.5cm}
    \caption{\method{} lets a coding agent direct the evolutionary search.
(a)~A fixed algorithm decides what the LLM sees in prior systems; in \method{}, the agent queries its memory itself.
(b)~The agent's loop: (1)~query the graph database, (2)~experiment, (3)~submit a candidate, (4)~receive its score and a steering message.
New sessions start with a fresh context and the same memory.
}
\vspace{-0.5cm}
\label{fig:teaser}
    
\end{figure}

We answer these questions affirmatively by developing \method{}, a novel framework for scientific discovery in which a coding agent both directs the evolutionary search and proposes its candidates, with every past attempt one query away in a graph database (Figure~\ref{fig:teaser}b, \S\ref{sec:methodology}, Table~\ref{tab:db-ablation}).
A coding agent works on the problem inside a sandbox, and its context is its working memory: it runs experiments, writes candidates, and decides what to try next.
Each agent session eventually ends and a fresh one takes over, for two reasons: a fresh agent is not conditioned on its predecessor's reasoning; and beyond some length, re-reading an ever longer context at every step costs more than starting over.
What a session learns outlives it in a long-term memory: a graph database of ideas and candidates.

Because the graph makes each candidate's ideas and lineage explicit, the selection rule of a classical algorithm becomes a single query: the best candidate per idea for MAP-Elites, or visit counts and values over the candidate lineage for Monte Carlo tree search.
The agent can use these rules, combine them, or replace them.
Because the search strategy is not fixed in advance, a stronger model improves not only the candidates but also the search that produces them, whereas under a fixed search algorithm it can improve only the candidates.
A server mediates all access to this memory, evaluates submitted candidates in isolation from the agent's sandbox, and returns each score with a message that keeps the agent on course within and across sessions.
A meta agent reads past sessions and writes skills that later sessions draw on.

We evaluate \method{} on tasks in kernel engineering, biology, algorithm design, and mathematics \citep{yuksekgonul2026learning} (Table~\ref{tab:ttt}), on five further AtCoder heuristic contests \citep{imajuku2025ale} (Table~\ref{tab:ahc}), on eleven mathematical and systems optimization tasks \citep{qu2026coral}, run with the same LLM as the published baselines (Table~\ref{tab:coral}), and on the kernel builder task \citep{anthropic2026takehome} (Figure~\ref{fig:kernel}).
In the course of this evaluation we found that several evaluators used by prior work admit reward hacking; we correct them and describe the common vulnerabilities.

In summary, our contributions are:
\begin{enumerate}
    \item We present \method{}, a framework for scientific discovery that replaces the fixed evolutionary search algorithm with a coding agent that directs the search and proposes candidates, and that manages its context efficiently through a database of past attempts in which the selection rules of classical search algorithms are single queries the agent can use, combine, or replace (\S\S\ref{subsec:prelim} and~\ref{subsec:framework}); reading past attempts from this database takes fewer LLM calls and costs less than reading them from plain files (Table~\ref{tab:db-ablation}), and the queries the agents write show which of these selection rules they actually use (Appendix~\ref{app:strategies}).
    \item We show that \method{} outperforms prior discovery frameworks on tasks in kernel engineering, biology, algorithm design, and mathematics, would have placed first against human competitors in seven past AtCoder contests, matches or improves the best published result on all of the eleven standard tasks, and improves the best published bound on the first autocorrelation inequality, while being more cost-efficient than the frameworks we compare against (\S\ref{sec:experiments}, Appendix~\ref{app:spend-curves}); its lead holds when runs are repeated (Appendix~\ref{app:repeats}).
    \item We identify reward-hacking vulnerabilities in evaluators used by prior work and correct them (Appendix~\ref{app:evaluators}).
\end{enumerate}

\section{Related Work}
\label{sec:related-work}

\textbf{LLM-based evolutionary search.}
A first line of work couples an LLM with an evolutionary search loop for
scientific and algorithmic discovery.
FunSearch \citep{romera2024mathematical} established the recipe, evolving
programs inside a fixed island-based loop and discovering new constructions in
extremal combinatorics. AlphaEvolve \citep{novikov2025alphaevolve} then shifted the landscape,
improving the best known results on roughly a fifth of more than fifty open
mathematical problems with a fixed archive and population policy combining
MAP-Elites and island models.
ShinkaEvolve keeps a prescribed population outer loop but adds novelty
filtering and bandit-based model selection, while MCTS-AHD replaces population
evolution with a fixed UCT-organized tree
\citep{lange2026shinkaevolve,zheng2025mctsahd}.
SimpleTES designs parallel
trajectories, refinement, and history selection,
and sets new best-known values with an open model \citep{ye2026simpletes}.
Fine-tuning variants keep the same loop: ThetaEvolve and TTT-Discover update
the model parameters themselves at test time
\citep{wang2025thetaevolve,yuksekgonul2026learning}.
In all these systems the search algorithm is designed in advance and stays
fixed; they differ mainly in which algorithm they fix.
A few go a step further and adapt the algorithm at runtime:
AdaEvolve adjusts within-island exploration, cross-island allocation, and
stagnation-triggered guidance through a prescribed three-level controller
\citep{cemri2026adaevolve};
\citet{gupta2026harness} run several fixed harnesses side by side and shift the
remaining compute to those whose early scores lead;
and EvoX evolves an explicit parent-selection, inspiration, and variation
policy, but only when a fixed progress monitor triggers replacement
\citep{liu2026evox}.
Even here, adaptation remains confined to a predefined space.

\textbf{Coding agents as proposers.}
A few systems instead hand the search to a coding agent.
AlgoEvo lets an agent decide when to inspect, diagnose, edit, or evaluate code,
but a fixed upper-confidence rule still chooses which earlier design guides
each step \citep{qiu2026algoevo}.
Vesper gives repository-level coding agents direct database access, but its
fixed island-model policy samples the parent and seeds the worktree before the
agent is invoked \citep{ishibashi2026vesper}.
Meta-Harness removes the selection rule entirely: a coding agent reads every
earlier harness with its scores and traces and decides what to change next,
though its object is the harness around a model rather than the solution to a
discovery task \citep{lee2026metaharness}.
CORAL also uses coding agents for scientific discovery \citep{qu2026coral}.
\method{}, however, places them in a completely different framework.
First, \method{} periodically resets the agent's context, which avoids paying
for an ever longer history (Figure~\ref{fig:reset}a) and helps the agent find new ideas (Figure~\ref{fig:reset}b--d).
Second, where CORAL shares attempts, notes, and skills as file-system artifacts,
\method{} keeps its long-term memory in a graph database of ideas and candidate
lineage. With it, a fresh agent needs fewer LLM calls to read past attempts (Table~\ref{tab:db-ablation}), and the agents express the selection rules of classical algorithms as single queries (Appendix~\ref{app:strategies}).
Third, \method{} lets the agent write its own tools and run experiments before
it submits a candidate, and an optional meta agent turns past sessions into
skills that help later sessions improve sooner and propose better candidates (Figure~\ref{fig:additional}c,d).
Finally, CORAL interrupts and restarts the session on a heartbeat, which wastes LLM calls (Appendix~\ref{app:steering}), whereas
\method{}'s server steers the running session with a message after every
submission.

\textbf{Adapting prompts, scaffolds, and memories.}
A third group adapts artifacts other than the candidate program.
With frozen weights, GEPA adapts prompts, BES searches over reasoning
trajectories at inference time, SICA recursively edits its coding-agent
scaffold, and SkillOpt applies bounded updates to skill documents
\citep{agrawal2025gepa,xu2026bidirectional,robeyns2025selfimproving,yang2026skillopt}.
Others persist semantic search state: SeaEvo clusters and retrieves strategy descriptions and PACEvolve decouples idea generation from code evaluation \citep{luo2026seaevo,yan2026pacevolve}.

\textbf{Autonomous research agents.}
A fast-growing line of work gives LLM agents autonomy over research itself.
The AI Scientist automates the cycle from idea to paper \citep{lu2024aiscientist,yamada2025aiscientistv2}; the AI co-scientist and the
Virtual Lab organize agents as a research group~\citep{gottweis2025coscientist,swanson2025virtuallab}; and
long-running systems combine literature search and data analysis to reach new
findings \citep{ghareeb2026multiagent,mitchener2025kosmos}. Agents are also
changing how research is shared: Paper2Agent turns papers into agents
\citep{miao2025paper2agent}, AgentRxiv lets agents build on each other's
reports \citep{schmidgall2025agentrxiv}, and concurrent projects run coding
agents autonomously on optimization targets
\citep{karpathy2026autoresearch,chen2026avo}. These systems target open-ended research judged by humans or experiments. AlphaEvolve-style methods instead perform verifier-driven program evolution, but the LLM only proposes candidates inside a fixed, human-designed search loop. \method{} combines the two: an autonomous agent that owns the search, scored by a verifier.

\vspace{-0.25cm}
\section{Methodology}
\label{sec:methodology}

\subsection{Problem Formulation and Motivation}
\label{subsec:prelim}

LLM-driven scientific discovery casts a research problem as a search over programs.
The problem is stated in natural language, and each \emph{candidate} is a program $x$: a representation the LLM can read and revise, and a machine can execute and score.
A fixed \emph{evaluator} $E$ scores each candidate with a scalar \emph{fitness} $E(x)$, oriented so that higher is better, and reports any auxiliary metrics the problem defines.
The goal is a candidate of high fitness within a budget of evaluations, found by an LLM that revises candidates iteratively in light of what has already been tried.

Existing systems \citep[e.g.,][]{romera2024mathematical,novikov2025alphaevolve,lange2026shinkaevolve,ye2024reevo} fix this iteration before the run: an evolutionary algorithm decides which past candidates the LLM sees, and the LLM only proposes the next one.
This design is ironic: these systems aim to surpass human design, yet their search procedure is human-designed, and the LLM, the one component meant to outdo human design, has no say in it.
Committing the whole run to one such procedure is hard to justify when no single strategy is best across problems (\S\ref{sec:introduction}).

We therefore replace the LLM proposer with a coding agent.
The agent queries past attempts to build its own context, runs its own experiments, and records what it tries in a memory that outlives its context.
Which candidates to revisit, which directions to abandon, and which strategy to follow become decisions the agent makes, and revises, during the run; in the terms of metacognition, it takes over the monitoring and control of its own search \citep{flavell1979metacognition}.
This flexibility pays off only if it remains safe and efficient over the long horizons these tasks demand; \method{} is the infrastructure that provides it.

\subsection{The \method{} Framework}
\label{subsec:framework}

\begin{figure}[t]
    \centering
    \includegraphics[width=\linewidth]{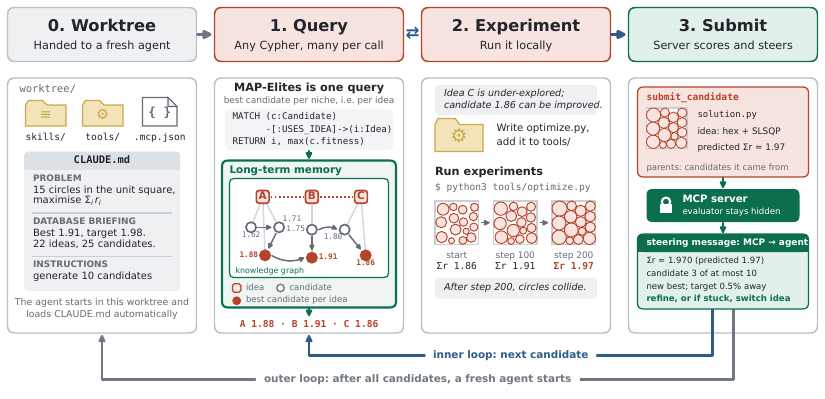}
    \vspace{-0.5cm}
    \caption{\method{} on a toy task: place 15 circles in a square to maximize
    the sum of their radii. A coding agent does the search. An MCP server
    stores all past ideas and candidates in a graph database and scores new
    candidates.
    (0)~The agent starts in its folder (worktree). A CLAUDE.md file there
    describes the task and a summary of the database.
    (1)~The agent explores the database with any read-only query.
    (2)~It runs experiments, using available tools or new ones it writes first.
    (3)~It submits a candidate. The server scores it with the task's evaluator
    and replies with a steering message.
    The agent repeats steps 1--3 for each new candidate (inner loop). After the
    last one, a new agent with a fresh context starts on the same database and
    keeps all tools written so far (outer loop).}
    \vspace{-0.5cm}
    \label{fig:walkthrough}
\end{figure}

\method{} has three components.
The \textbf{search agent} is a coding agent that studies previous attempts, plans, runs experiments, and proposes new candidates.
The \textbf{long-term memory} is a graph database that keeps every idea and candidate tried so far.
The optional \textbf{meta agent} is another coding agent that reflects on the search agents' reasoning logs and writes what it learns as skills for them.
We designed these components with two properties in mind, and show below how they achieve them.
First, \emph{flexibility}: no search strategy is hard-coded, and the agent decides what to read, what to try, and when to change course.
Second, \emph{efficiency over long horizons}: a search runs for hours, so the agent must stay on course, retain what it learns beyond its context, and keep its token cost under control.

\textbf{Search agent:} The search agent is a coding-agent CLI, such as Claude Code, sandboxed in a version-controlled worktree.
It starts from an instruction file that describes the task, summarizes the database, and explains how to access it through a Model Context Protocol (MCP) server; the worktree also holds tools and skills folders, both empty in the first session.
At each step, the agent chooses whether to query the database, plan, write or run tools, or submit a candidate, and it reflects on each evaluation and plans further.
When the agent finds that it has stalled, or that carrying its context costs more than starting afresh, it records a lesson in the database, including experiments that never became candidates, and terminates, ending the \emph{session}.
A new session then starts a fresh search agent that inherits the long-term memory, tools, and skills. Figure~\ref{fig:walkthrough} walks through a session on a toy task.

\textbf{Long-term memory:} The long-term memory is a graph database of all ideas, candidates, sessions, and lessons, and the relations among them.
It sits on the MCP server alongside the evaluator: the agent reads it with any query but writes only through MCP tools that validate their input arguments, keeping the database consistent.
The server scores each submitted candidate and sends the agent a steering message that reports the fitness and the number of candidates left, keeping the agent on course over the long horizon.
The graph also makes classical strategies native: MAP-Elites, which keeps the best candidate per niche, is one query returning the best candidate for each idea.

\textbf{Meta agent:}
The meta agent is another coding-agent CLI, but with a different task and a different view of the database.
It runs after each search-agent session and focuses on how the search was done rather than on what it found.
It reads the session transcripts, including the search agents' reasoning traces, and relates them to the outcomes recorded in the long-term memory to find which ways of reasoning paid off and which did not.
It then consolidates its conclusions into skill files that later sessions draw on.

The agent accumulates a long context, which causes three problems.
(i)~Each LLM call re-reads the full context, so past some length a new session is cheaper than continuing, even with the reduced cost of cached tokens.
(ii)~The agent also becomes anchored to its earlier reasoning, which makes it harder to change ideas and more likely to get stuck in a local optimum.
(iii)~It is more likely to lose track of the task.
Fresh agents and the database together address the first two problems.
The new agent explores the database from scratch and can form a new perspective on the search; we expect this to produce new ideas more reliably than prompting the old agent to try something else while its context stays the same.
Restarting is cheap because retrieval is a query (Table~\ref{tab:db-ablation}).
Without the database, the agent would read past attempts from files: the Meta-Harness \citep{lee2026metaharness}, for example, reads a median of 82 files per iteration, each read costing an LLM call and consuming context the agent cannot get back.
For the third problem, the server answers every submission with a steering message that tells the agent where the search stands and keeps it on course.

\vspace{-0.2cm}
\section{Experiments}
\label{sec:experiments}

We evaluate \method{} on 24 problems spanning five systems tasks, six mathematical optimization tasks, seven AtCoder heuristic contests, the first and second autocorrelation inequalities (AC1 and AC2), two GPU kernel engineering tasks, single-cell denoising, and Anthropic's kernel builder task. We compare solution quality with existing search frameworks and published test-time finetuning results (Tables~\ref{tab:ttt}--\ref{tab:kernel}), including comparisons at matched model spend (Figure~\ref{fig:kernel}, Appendix~\ref{app:spend-curves}). We also study the effects of context resets (Figure~\ref{fig:reset}), parallel search agents (Figure~\ref{fig:additional}a,b), meta-agent guidance (Figure~\ref{fig:additional}c,d), and the graph database (Table~\ref{tab:db-ablation}). We provide hyperparameters and run configurations in Appendix~\ref{app:details}.

\paragraph{Benchmarks and baselines.}
We organize the evaluation into four settings:
TTT-Discover's multi-domain discovery tasks
\citep{yuksekgonul2026learning}, AtCoder heuristic
contests, systems and mathematical optimization, and
Anthropic kernel builder
(Tables~\ref{tab:ttt}--\ref{tab:kernel}).

\textbf{Multi-domain discovery tasks.}
We first evaluate on tasks from four domains: kernel
engineering, biology, algorithm design, and mathematics.
The tasks are taken from TTT-Discover
\citep{yuksekgonul2026learning}, which reports results obtained
by finetuning gpt-oss-120b with reinforcement learning
(Table~\ref{tab:ttt}). We run both \method{} and CORAL on
each task with the same LLM and the same budget: GPT-6 Sol
with \$10 per task for kernel engineering and biology, and
Claude Opus~5 with \$100 per task for algorithm design and
mathematics. Web search is enabled for both methods on AC1
and AC2, the only experiments in this paper that use it.

\begin{table}[!ht]
\centering
\small
\caption{Results on kernel engineering, biology, algorithm design, and
mathematics tasks. CORAL and \method{} run at matched model spend: GPT-6 Sol
at \$10 per task on the first two categories, Claude Opus~5 at \$100 on the last
two; TTT-Discover results are from \citet{yuksekgonul2026learning}. Kernel
entries are runtimes in $\mu$s (TriMul on A100, MLA-Decode on H200); biology
entries are single-cell denoising MSE. Best in bold.
$^{\dagger}$Published only on AMD MI300X (1{,}669--1{,}706~$\mu$s).
$^{\ddagger}$Published program re-scored by our evaluator on the search split.}
\label{tab:ttt}
\label{tab:ttt-tasks}
\setlength{\tabcolsep}{8pt}
\renewcommand{\arraystretch}{1.25}
\begin{tabular}{llccc}
\toprule
& Task & TTT-Discover & CORAL & \method{} \\
\midrule
\multirow{2}{*}{\rotatebox[origin=c]{90}{Kernel}}
& TriMul $\downarrow$      & 2{,}198 & 2{,}419 & \textbf{2{,}080} \\
& MLA-Decode $\downarrow$  & --$^{\dagger}$ & 1{,}582 & \textbf{1{,}474} \\
\midrule
\multirow{3}{*}{\rotatebox[origin=c]{90}{Biology}}
& Pancreas $\downarrow$    & 0.1655$^{\ddagger}$ & 0.1676 & \textbf{0.1603} \\
& PBMC $\downarrow$        & 0.154 & 0.156 & \textbf{0.147} \\
& Tabula lung $\downarrow$ & 0.136 & 0.139 & \textbf{0.132} \\
\midrule
\multirow{2}{*}{\rotatebox[origin=c]{90}{Algo.}}
& AHC039 $\uparrow$        & 567,062     & 566,654     & \textbf{577,217} \\
& AHC058 $\uparrow$        & 848,414,228 & 846,389,623 & \textbf{848,724,866} \\
\midrule
\multirow{2}{*}{\rotatebox[origin=c]{90}{Math}}
& AC1 $\downarrow$         & 1.50287     & 1.50282     & \textbf{1.50274} \\
& AC2 $\uparrow$           & 0.9591      & 0.95755     & \textbf{0.96275} \\
\bottomrule
\end{tabular}
\end{table}

\textbf{AtCoder heuristic contests.}
To assess performance on heuristic programming problems
with official tests and human competition results, we
evaluate on five past AtCoder Heuristic Contests:
AHC006, AHC012, AHC015, AHC025, and AHC041
(Table~\ref{tab:ahc}). These contests provide the task
setting used by ALE-Bench \citep{imajuku2025ale}.
We run OpenEvolve, EvoX, CORAL, and \method{} with
Claude Opus~5 and score their programs on the official
system tests.

\begin{table}[!ht]
\centering
\scriptsize
\setlength{\tabcolsep}{1.3pt}
\caption{Results on past AtCoder heuristic contests with Claude Opus~5, scored on the official system tests. Methods are compared at equal model spend on each contest (\$120--\$200, depending on the contest). \emph{Con.} is the rank among the original competitors; \emph{Ext.} includes post-contest submissions.
$^{\dagger}$Absolute score, lower is better; ranks follow the official relative-score standings; see Appendix~\ref{app:ahc025-scoring}.}
\label{tab:ahc}
\setlength{\tabcolsep}{3pt}
\resizebox{\textwidth}{!}{%
\begin{tabular}{l ccc ccc ccc ccc}
\toprule
& \multicolumn{3}{c}{OpenEvolve} & \multicolumn{3}{c}{EvoX}
& \multicolumn{3}{c}{CORAL} & \multicolumn{3}{c}{\method{}} \\
\cmidrule(lr){2-4} \cmidrule(lr){5-7} \cmidrule(lr){8-10} \cmidrule(lr){11-13}
Task & Score & Con. & Ext. & Score & Con. & Ext. & Score & Con. & Ext. & Score & Con. & Ext. \\
\midrule
AHC006 $\uparrow$ & 2,238,362 & 1st & 13th & 2,197,920 & 4th & 36th
  & 2,317,352 & 1st & 1st & \textbf{2,324,852} & 1st & 1st \\
AHC012 $\uparrow$ & 99,090,668 & 5th & 39th & 98,579,333 & 9th & 55th
  & 99,977,717 & 1st & 4th & \textbf{99,997,290} & 1st & 3rd \\
AHC015 $\uparrow$ & 144,369,058 & 15th & 130th & 151,853,188 & 4th & 93rd
  & 156,230,796 & 4th & 57th & \textbf{168,660,525} & 1st & 5th \\
AHC025$^{\dagger}$ $\downarrow$ & 95,926,386 & 37th & 51st & 39,086,466 & 21st & 30th
  & 25,923,844 & 2nd & 6th & \textbf{19,173,103} & 1st & 3rd \\
AHC041 $\uparrow$ & 76,321,331 & 10th & 58th & 75,787,437 & 31st & 127th
  & 77,358,800 & 1st & 6th & \textbf{78,044,915} & 1st & 1st \\
\bottomrule
\vspace{-0.2cm}
\end{tabular}}
\vspace{-0.2cm}
\end{table}

\textbf{Systems and mathematical optimization.}
To test performance across different optimization problems,
we consider five systems tasks and six mathematical tasks
(Table~\ref{tab:coral}). We compare against OpenEvolve
\citep{openevolve}, ShinkaEvolve
\citep{lange2026shinkaevolve}, EvoX \citep{liu2026evox},
and CORAL \citep{qu2026coral}, using the baseline results
reported by \citet{qu2026coral}, except on signal processing,
where we rerun all baselines with the corrected evaluator. We run \method{} with Claude
Opus~4.6, the model used for those reported results, for a
fair comparison.

\begin{table}[!ht]
\centering
\small
\setlength{\tabcolsep}{2.7pt}

\caption{Results on five systems and six mathematical
optimization tasks. All methods use Claude Opus~4.6 with
a budget of at most 100 candidates. Published baseline
results are taken from \citet{qu2026coral}.
Arrows indicate the direction of improvement; best scores
and ties at common reporting precision are bold.
For signal processing, the faded row reports scores from
the original evaluator, which permits reward hacking;
these scores are excluded from performance comparisons.
The next row uses the corrected evaluator
(Appendix~\ref{app:evaluators}).}
\label{tab:coral}
\begin{tabular}{llccccc}
\toprule
& Task & OpenEvolve & ShinkaEvolve & EvoX & CORAL & \method{} \\
\midrule
\multirow{5}{*}{\rotatebox[origin=c]{90}{Systems}}
& EPLB $\uparrow$         & 0.127   & 0.129   & 0.146   & 0.149  & \textbf{0.153} \\
& PRISM $\uparrow$         & \textbf{26.26} & \textbf{26.26} & \textbf{26.26} & \textbf{26.26} & \textbf{26.26} \\
& LLM-SQL $\uparrow$       & 0.716   & 0.724   & 0.726   & 0.731   & \textbf{0.737} \\
& Txn Sched.\ $\uparrow$   & 3774    & 3802    & 3984    & 4566    & \textbf{4651} \\
& Cloudcast $\downarrow$   & 627.2   & 627.8   & 623.5   & 618.4   & \textbf{618.0} \\
\midrule
\multirow{7}{*}{\rotatebox[origin=c]{90}{Math}}
& Circle-Pack.\ $\uparrow$ & 2.6293  & 2.6001  & 2.6320  & \textbf{2.6360} & \textbf{2.6360} \\
& \faded{Signal Proc.\ $\uparrow$ (original)} & \faded{0.6420} & \faded{0.8171} & \faded{0.7306} & \faded{0.8229} & \faded{\textbf{0.9259}} \\
& Signal Proc.\ $\uparrow$ (corrected) & 0.7286 & 0.7635 & 0.7288 & 0.7790 & \textbf{0.7948} \\
& Erd\H{o}s Over.\ $\downarrow$ & 0.38188 & 0.38156 & 0.38125 & {0.38089} & \textbf{0.38077} \\
& MMD-16-2 $\downarrow$    & 12.92   & \textbf{12.89} & 12.96 & \textbf{12.89} & \textbf{12.89} \\
& MMD-14-3 $\downarrow$    & 4.21    & 4.46    & 4.46    & \textbf{4.16} & \textbf{4.16} \\
& 3rd-Autocorr.\ $\downarrow$ & 1.4731 & 1.4812 & 1.5552 & 1.4557 & \textbf{1.4556} \\
\bottomrule
\end{tabular}
\end{table}

\textbf{Anthropic kernel builder.}
Finally, we test how solution quality improves with model
spend on Anthropic's Original Performance Take-Home
\citep{anthropic2026takehome}
(Table~\ref{tab:kernel}). The task is to optimize a
parallel tree-traversal kernel on a simulated VLIW SIMD
processor. At each step, the kernel hashes an input value
with the current node's value and uses the result to
choose the next branch. The goal is to minimize execution
cycles while preserving correctness. We compare \method{}
with all four search baselines at common model-spend
budgets, using Claude Opus~5 with low reasoning effort
and the same evaluator.

\begin{table}[!ht]
\centering
\small
\setlength{\tabcolsep}{5pt}
\caption{Kernel builder at common model-spend budgets.
All methods use Claude Opus~5 with low reasoning effort and
the same evaluator. Each entry is the lowest cycle count
achieved at or below the indicated budget; lower is better.
The best result at each budget is bold.}
\label{tab:kernel}
\begin{tabular}{rccccc}
\toprule
Budget & OpenEvolve & ShinkaEvolve & EvoX & CORAL & \method{} \\
\midrule
\$25  & 1,248 & 1,344 & 1,225 & 1,257 & \textbf{1,016} \\
\$50  & 1,213 & 1,218 & 1,153 & 1,214 & \textbf{976} \\
\$75  & 1,178 & 1,210 & 1,119 & 1,157 & \textbf{973} \\
\$100 & 1,174 & 1,187 & 1,098 & 1,148 & \textbf{960} \\
\$118 & 1,174 & 1,156 & 1,098 & 1,148 & \textbf{960} \\
\bottomrule
\end{tabular}
\vspace{-0.2cm}
\end{table}

\vspace{-0.2cm}
\paragraph{Protocol.}
For the AtCoder contests (Table~\ref{tab:ahc}), all methods
share the same model-spend budget on each contest. For the
systems and mathematical tasks (Table~\ref{tab:coral}),
\citet{qu2026coral} report no model spend, only that each
method ran for 100 candidates; we therefore allow at most 100
submitted candidates per run, although \method{} converges
well before that. Where model spend is available, \method{}
is more cost-efficient than the other frameworks
(Appendix~\ref{app:spend-curves}). For kernel
optimization (Table~\ref{tab:kernel}), we compare all
methods at common model-spend budgets from \$25 to \$118.
Each main-table entry reports the best candidate found
within the stated budget in a single run. Over repeated runs on nine tasks, \method{} has the better mean on all of them, and on five its worst run beats every baseline's best (Appendix~\ref{app:repeats}).
Appendix~\ref{detail_spend} explains how model spend is
estimated from token usage and model-specific prices.

We use evaluators from prior work and correct vulnerabilities
that allow candidates to obtain high scores without solving
the intended task. Where we report scores from both the
original and corrected evaluators, we mark the original
scores separately and exclude results that exploit these
vulnerabilities from performance claims. The corrections
preserve the scoring of valid solutions.
Appendix~\ref{app:evaluators} describes the vulnerabilities
and fixes.

\subsection{Results}
\label{sec:results}

\paragraph{Multi-domain discovery tasks.}
\method{} outperforms both TTT-Discover and CORAL on every task in Table~\ref{tab:ttt}. On AHC039 and AHC058, its programs would have placed first in
both original contests. Its AC1 upper bound of
$1.50274$ ranks first on the public Einstein Arena leaderboard \citep{bianchi2026einsteinarena} at the time of writing (Appendix~\ref{app:ahc-budgets}).

\vspace{-0.25cm}
\paragraph{AtCoder heuristic contests.}
\method{} achieves scores that would have placed first in all five original AtCoder contests (Table~\ref{tab:ahc}). It also outperforms the compared search frameworks on all five tasks. When post-contest submissions are
included, \method{} remains in the top five on every task, ranking first on two. For AHC025, we report the fixed absolute score and provide the relative scores of the official standings in Appendix~\ref{app:ahc-budgets}.

\vspace{-0.25cm}
\paragraph{Systems and mathematical optimization.}
\method{} matches or outperforms every baseline on all
eleven tasks in Table~\ref{tab:coral}, leading on seven
and tying for best on four at common reporting precision.
It achieves the best score on four of the five systems
tasks: EPLB, LLM-SQL, transaction scheduling, and
Cloudcast; and it reaches the known optimum on PRISM.
On the mathematical tasks, it achieves the best score
under the corrected signal-processing evaluator and
improves on all baselines for both Erd\H{o}s overlap
and the third autocorrelation inequality. It matches
the strongest results on circle packing, MMD-16-2,
and MMD-14-3. These gains are obtained using the same
backbone model across all methods.

\vspace{-0.25cm}
\paragraph{Anthropic kernel builder.}
\method{} achieves the lowest cycle count at every reported budget (Table~\ref{tab:kernel} and Figure~\ref{fig:kernel}). With just \$25 of model spend, it reaches 1,016 cycles, outperforming every baseline even at their full \$118 budget. By \$100, it reaches
960 cycles, compared with 1,098 for the strongest baseline, EvoX, a 12.6\% reduction in execution cycles. This corresponds to a $153.9\times$ speedup over the naive kernel. Appendix~\ref{app:spend-curves} shows the same pattern of performance against model spend on more tasks.

\subsection{Additional Experiments}
\label{sec:additional}

\begin{wrapfigure}{r}{0.55\linewidth}
  \centering
  \vspace{-12pt}
  \includegraphics[width=\linewidth]{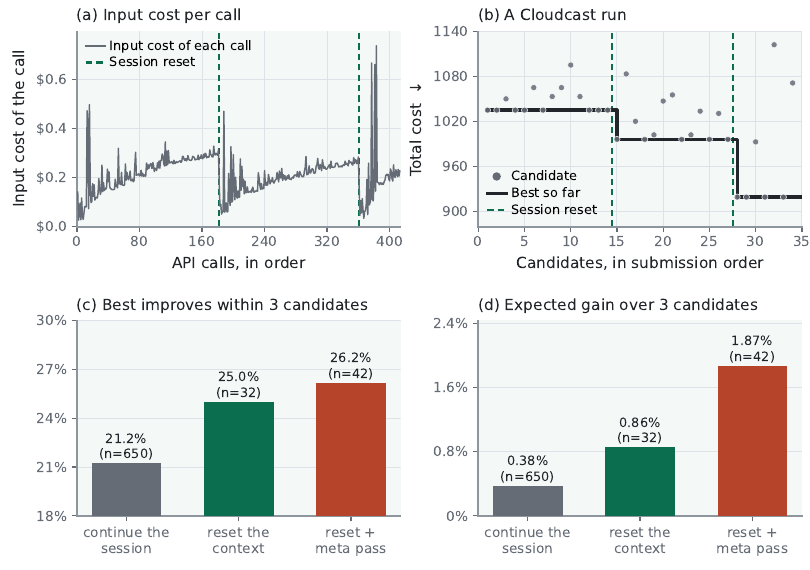}
  \vspace{-14pt}
  \caption{Context resets. (a) Without a reset, the input cost per call
    keeps growing. (b) After a reset, the agent
    sees the search afresh, making new ideas more likely. (c,d) After a stall
    of over five candidates, a reset raises the chance and size of an
    improvement; a meta pass raises them further.}
  \label{fig:reset}
  \vspace{-16pt}
\end{wrapfigure}
We study four components of \method{}:
context resets, parallel search agents, meta-agent guidance, and the graph database.

\textbf{Context resets.}
\method{} resets the context when the agent realizes that it has stalled, or
when carrying a long cached context costs more than the input tokens of a fresh
one. This helps in two ways (Figure~\ref{fig:reset}). First, without a reset,
the cost of each call keeps increasing (Figure~\ref{fig:reset}a). Second, after
a reset the agent explores again, gains a new perspective, and is more likely
to escape a local optimum, as in the Cloudcast run of
Figure~\ref{fig:reset}b, where every improvement comes right after a reset.
Figures~\ref{fig:reset}c and~\ref{fig:reset}d show that this holds
on average across all our experiments: when the search has stalled for more
than five candidates, a reset raises both the chance of improving the best
score and the size of the improvement. The meta agent raises both further.

\begin{figure}[t]
  \centering
  \begin{minipage}[t]{0.245\linewidth}
    \centering
    \includegraphics[width=\linewidth]{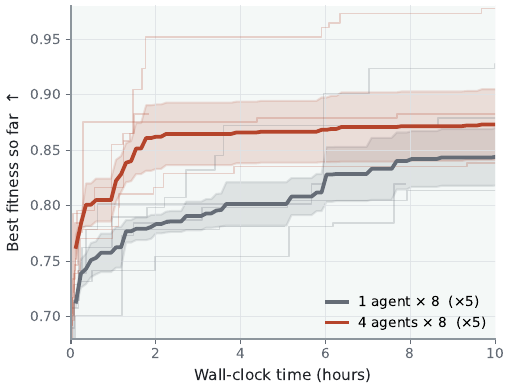}\\[-2pt]
    {\scriptsize (a) Parallel agents, time}
  \end{minipage}\hfill
  \begin{minipage}[t]{0.245\linewidth}
    \centering
    \includegraphics[width=\linewidth]{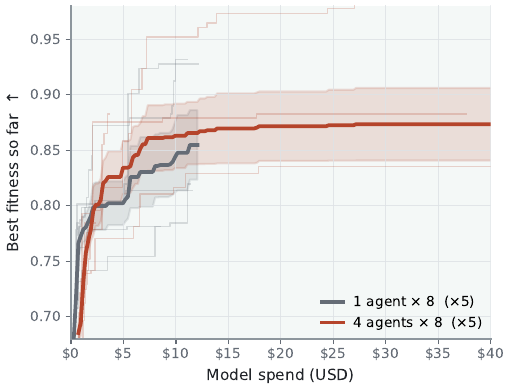}\\[-2pt]
    {\scriptsize (b) Parallel agents, spend}
  \end{minipage}\hfill
  \begin{minipage}[t]{0.245\linewidth}
    \centering
    \includegraphics[width=\linewidth]{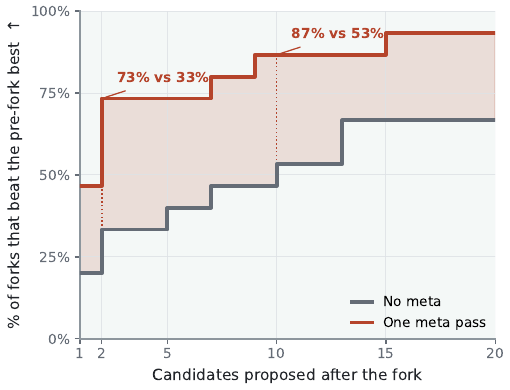}\\[-2pt]
    {\scriptsize (c) Meta agent, how soon it improves}
  \end{minipage}\hfill
  \begin{minipage}[t]{0.245\linewidth}
    \centering
    \includegraphics[width=\linewidth]{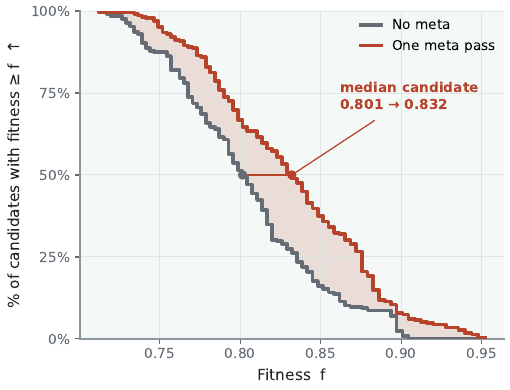}\\[-2pt]
    {\scriptsize (d) Meta agent, all candidates' fitness}
  \end{minipage}
  \vspace{-4pt}
  \caption{(a,b)~Four parallel agents find good solutions sooner in wall-clock time
    than one agent, but not at lower model spend. Lines show the mean over
    five runs, bands the standard error, and faint lines individual runs.
    (c)~With meta guidance, a new search agent beats the best
    score so far sooner: within two candidates in 73\% of cases, against 33\%
    without meta guidance. (d)~With meta guidance, the candidates the agent proposes are better
    overall, not only its best one: each curve shows the share of candidates
    that reach at least a given fitness, and the median candidate rises from
    0.801 to 0.832.}
  \label{fig:additional}
  \vspace{-10pt}
\end{figure}

\textbf{Parallel search agents.}
\method{} also supports several search agents in parallel, all
reading from and writing to the same candidate database
through one MCP server. We compare one search agent with four
on transaction scheduling with Claude Haiku~4.5,
using five runs per setting. With four agents, the search
converges faster in wall-clock time
(Figure~\ref{fig:additional}a), since the agents propose and
evaluate up to four times as many candidates in the same time.
At equal model spend, however, we do not see a clear
difference between the two settings
(Figure~\ref{fig:additional}b). Thus, parallel agents
shorten the time to a good solution without reducing its cost,
which makes them useful when wall-clock time matters more than
model spend.

\textbf{Meta-agent guidance.}
To isolate the effect of the meta agent, we run
transaction scheduling five times without a meta agent.
We then fork each run three times, after its first, second, and third
iteration. A fork copies the run's database and workspace at that point,
adds a single meta pass, and continues for two more iterations with no
further guidance; the original run continues from the same point without
guidance. A fork and its original therefore start from the same state, and
the only difference between them is whether the next search agent has read
the meta agent's guidance. This removes most of the run-to-run randomness
from the comparison.
Across the 15 forks, guidance helps the new agent make progress sooner: it
beats the best fitness found before the fork within its first two
candidates in 73\% of the forks, against 33\% without guidance, and within
ten candidates in 87\% against 53\% (Figure~\ref{fig:additional}c).
Guidance also raises the quality of what the agent proposes, not only of its
best candidate: across all candidates proposed in the two iterations after
the fork, the median fitness rises from 0.801 to 0.832
(Figure~\ref{fig:additional}d). At the end of these two iterations, the
guided run has the higher best fitness in 10 of the 15 forks, the same in 2,
and a lower one in 3.
We use Claude~Haiku~4.5 as the search agent and Claude~Opus~4.6 as the meta
agent, with ten candidates per iteration and five iterations per run.

\begin{wraptable}{r}{0.5\linewidth}
\vspace{-10pt}
\centering\small
\setlength{\tabcolsep}{4pt}
\begin{tabular}{lrr}
\toprule
before the first candidate & graph database & plain files \\
\midrule
LLM calls & $\mathbf{28 \pm 7}$ & $38 \pm 8$ \\
tokens read (millions) & $\mathbf{2.8 \pm 0.8}$ & $3.6 \pm 1.0$ \\
price (\$) & $\mathbf{4.1 \pm 0.9}$ & $5.1 \pm 1.0$ \\
\bottomrule
\end{tabular}
\caption{Graph database vs.\ plain files, same eight sessions: the database
agent reads its history in fewer LLM calls and at lower cost. Mean $\pm$
standard error.}
\label{tab:db-ablation}
\vspace{-8pt}
\end{wraptable}
\textbf{Graph database vs.\ plain files.}
To isolate the effect of the database, we replay eight sessions with the
history stored in a file system instead. For each session we rebuild the
state it started from and run it again with one difference: the agent reads
the same history from plain files, laid out as in CORAL. Model,
instructions, budget and history are unchanged. We compare the two settings
up to the first candidate each session submits. This is the exploration
phase: a fresh session does most of its reading of the history at the start,
before it has anything to submit. On average this phase takes a quarter of a
session's LLM calls but holds two thirds of its history look-ups.
Table~\ref{tab:db-ablation} compares the two settings on the LLM calls,
tokens and cost spent in this phase. With files, the agent makes $1.4\times$
as many LLM calls and reads $1.3\times$ as many tokens, yet pulls less
history text into its context: one query returns what files yield only over
several calls. In every file session the agent writes a script that opens
and sorts all records to rebuild the leaderboard, which the database returns
with one query. With the database, three of the eight sessions queried the
best candidate per idea, the MAP-Elites selection, in one call.

\FloatBarrier
\section{Conclusion}
We asked whether a search strategy chosen by a human before the run scales
better than promoting the model from proposer to planner and letting it own
the search.
Our results favor the agent.
The agent reached better scores at lower cost than the frameworks we
compared against (Tables~\ref{tab:ttt}--\ref{tab:kernel}, Figure~\ref{fig:kernel}), would have placed first among human competitors in seven
past AtCoder contests (Tables~\ref{tab:ttt} and~\ref{tab:ahc}), and improved the best published bound
on the first autocorrelation inequality (Table~\ref{tab:ttt}).
The improvement over fixed search algorithms comes, in our view, from the
agent seeing the whole search rather than the few candidates a fixed algorithm
would hand it, and from reasoning and experimenting over many steps rather
than proposing each candidate in one shot.
The improvement over CORAL, where an agent also runs the search, comes from
managing context more efficiently over the long horizon.
We showed that resetting the agent's context kept the cost of each call low and
made it more likely to escape a stall (Figure~\ref{fig:reset}), and that a meta agent helped further (Figure~\ref{fig:additional}c,d).
A graph database let each fresh agent re-explore the search cheaply (Table~\ref{tab:db-ablation}) and express classical selection rules as single queries (Appendix~\ref{app:strategies}), while the agent wrote its own
tools and ran experiments before submitting.
Together, these made \method{} more cost-efficient than CORAL and other frameworks we
compared against (see Appendix~\ref{app:spend-curves} for more tasks and figures).
Along the way, we found that several evaluators inherited from prior work
admit reward hacking, and corrected them (Appendix~\ref{app:evaluators}).

\textbf{Limitations.}
\method{} gives an agent an efficient environment to work on a scientific discovery
problem for many hours, but it cannot go beyond the capabilities of the underlying LLM.
It also applies only to problems with a quantitative evaluator that can score each candidate
automatically.

\section{Acknowledgments}

Portions of this research were conducted with the advanced computing resources provided by Texas A\&M High Performance Research Computing. We thank the Department of Electrical and Computer Engineering (ECEN) at Texas A\&M University for providing access to the Olympus computing cluster, and the Texas A\&M University System for providing access to the VISION supercomputer. This work was supported, in part, by National Science Foundation (NSF) under grants CNS-2148354, FuSe2-2425399, U.S. Army Combat Capabilities Development Command (DEVCOM) under Grant Number W911NF2520046. This work was also supported in part by the National Science Foundation grant NSF-CNS 2148354, federal agencies, and industry partners as specified in the Resilient \& Intelligent NextG Systems (RINGS) program. Any opinions, findings, and conclusions or recommendations expressed in this material are those of the authors and do not necessarily reflect the views of the sponsoring agencies.

\newpage

\subsection*{AI Use Statement}
Coding agents built on Claude and GPT models are the object of study in this
paper: they act as the search and meta agents of \method{} and as the
backbone of every baseline we run. That use is part of the method and is
described in Section~\ref{sec:methodology} and Appendix~\ref{app:details}. 

Separately from the method itself, we used generative AI tools
(Claude Code and ChatGPT) to help implement methods, namely to write and debug parts of the \method{} codebase and the analysis and plotting scripts.
We have not used generative AI tools to generate synthetic datasets,
develop conceptual frameworks, formulate mathematical claims or proofs,
propose hypotheses, design the methodology or experiments, or interpret
results; translation, dataset cleaning, and qualitative data analysis are
not applicable to this work.
Additionally, we used generative AI tools to search for and summarize
related literature, and to edit the paper for readability.
We have reviewed all AI-assisted work. All AI-assisted code was read and
tested by the authors, every reported number was checked against the
stored run logs and evaluator outputs, and every citation was checked
against the original source.
We take responsibility for the final content of this work, including
text, claims, or artifacts produced with the aid of generative AI.

\clearpage

\bibliography{references}
\bibliographystyle{iclr2027_conference}

\appendix

\clearpage
\section{Spend--Performance Curves}
\label{app:spend-curves}

\begin{figure}[!htbp]
  \setlength{\abovecaptionskip}{3pt}
  \centering
  \begin{minipage}[t]{0.49\textwidth}
    \centering
    \includegraphics[width=\linewidth]{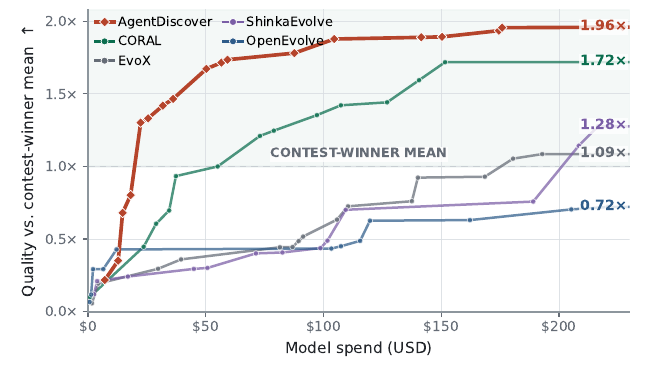}
    \textbf{(a) AHC025}
  \end{minipage}\hfill
  \begin{minipage}[t]{0.49\textwidth}
    \centering
    \includegraphics[width=\linewidth]{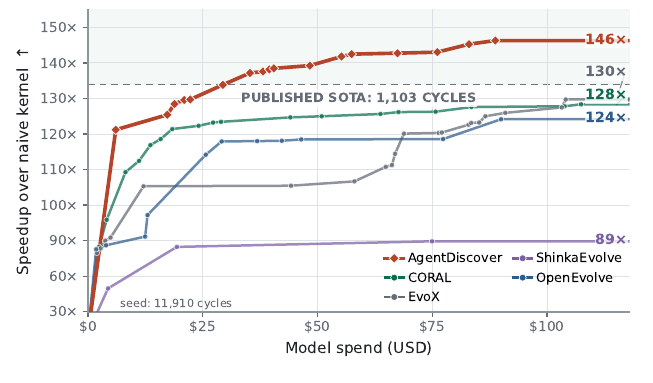}
    \textbf{(b) Kernel builder, second run}
  \end{minipage}
  \caption{(a) AHC025 fixed-objective comparison at matched model spend. The trajectories
  compare methods by their best mean absolute objective
  on the same 50 public evaluation cases, shown as the contest winner's mean divided by
  it (higher is better).
  The dashed $1\times$ reference uses the contest winner's mean over the 5,000
  official system-test cases, whereas the trajectories use the 50 public cases.
  Official leaderboard (relative) scores are reported in Table~\ref{tab:ahc025-absolute}.
  (b) Second run of Figure~\ref{fig:kernel}.}
  \label{fig:ahc025-budget}
\end{figure}

\begin{figure}[!htbp]
  \setlength{\abovecaptionskip}{3pt}
  \centering
  \begin{minipage}[t]{0.32\textwidth}
    \centering
    \includegraphics[width=\linewidth]{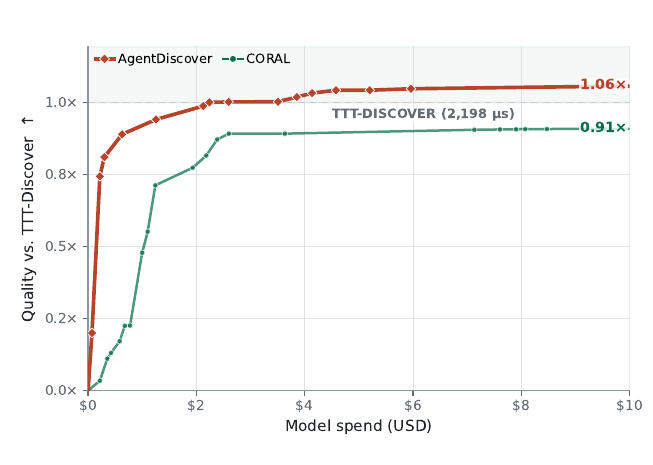}
    \textbf{(a) TriMul}
  \end{minipage}\hfill
  \begin{minipage}[t]{0.32\textwidth}
    \centering
    \includegraphics[width=\linewidth]{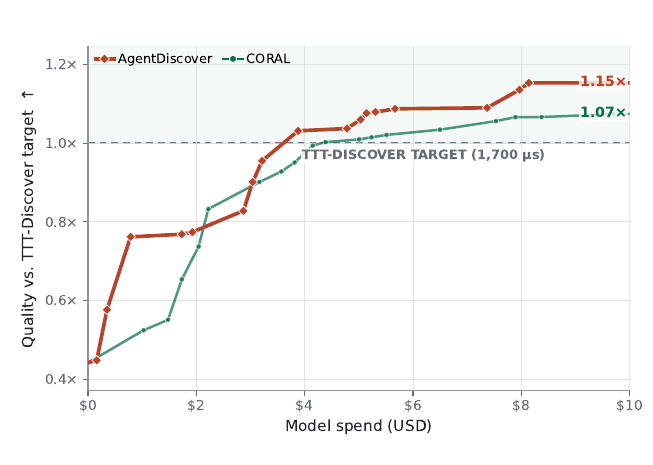}
    \textbf{(b) MLA-Decode}
  \end{minipage}\hfill
  \begin{minipage}[t]{0.32\textwidth}
    \centering
    \includegraphics[width=\linewidth]{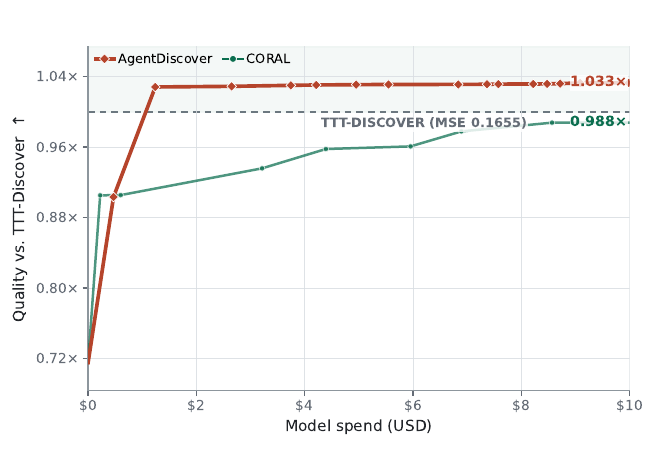}
    \textbf{(c) Denoising}
  \end{minipage}
  \caption{Best-so-far quality against model spend on the three TTT-Discover
  tasks, \method{} and CORAL with GPT-6 Sol, cut at the \$10 budget of
  Table~\ref{tab:ttt-tasks}. Quality is the TTT-Discover reference divided by
  the candidate's objective (runtime for the kernels, pancreas MSE for
  denoising), so higher is better and $1\times$ matches TTT-Discover; for
  MLA-Decode the reference is the 1,700~$\mu$s target of its prompt. \method{}
  reaches each reference sooner and ends higher on all three tasks.}
  \label{fig:ttt-tasks}
\end{figure}

\begin{figure}[!htbp]
  \setlength{\abovecaptionskip}{3pt}
  \centering
  \begin{minipage}[t]{0.49\textwidth}
    \centering
    \includegraphics[width=\linewidth]{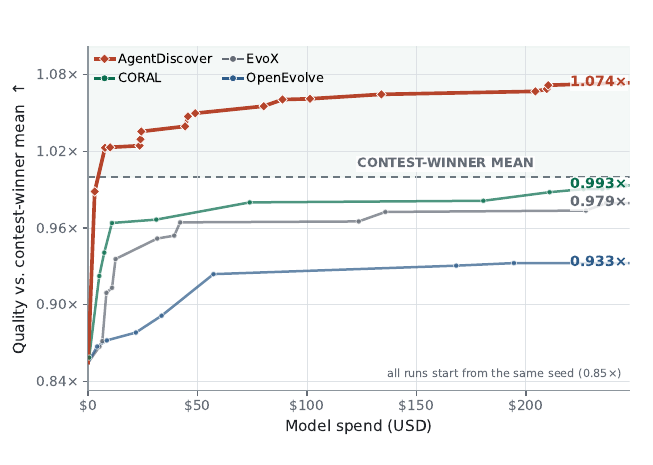}
    \textbf{(a) AHC015}
  \end{minipage}\hfill
  \begin{minipage}[t]{0.49\textwidth}
    \centering
    \includegraphics[width=\linewidth]{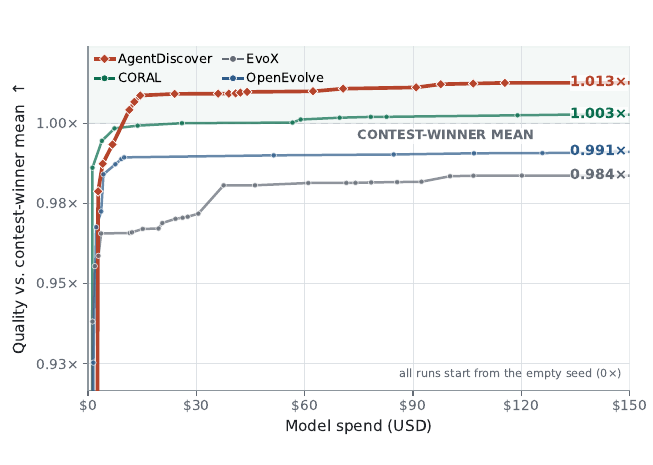}
    \textbf{(b) AHC041}
  \end{minipage}
  \caption{Best-so-far performance against billed model spend on two further
  AtCoder heuristic contests, measured on the public evaluation cases. Each
  objective is normalized by the corresponding contest-winner mean, so higher is
  better and $1\times$ is the reference value.}
  \label{fig:ahc-standalones}
\end{figure}

This section reports best-so-far performance against model spend for the tasks whose
main-text tables report only final scores.
Figure~\ref{fig:ahc025-budget} compares all five methods on the AHC025 absolute
score (Appendix~\ref{app:ahc025-scoring}) as a function of model spend. \method{} reaches the contest-winner
reference after \$22.34 of model spend, against \$72.97 for the strongest baseline.
At the full budget, \method{} finishes at $1.96\times$ the contest-winner quality
against CORAL's $1.72\times$, a 12.3\% lower objective.
In a second run of the kernel builder (Figure~\ref{fig:ahc025-budget}b), \method{}
again finishes best: 1,010 cycles against EvoX's 1,139.
Figure~\ref{fig:ttt-tasks} shows the same curves for TriMul, MLA-Decode, and denoising.
Figure~\ref{fig:ahc-standalones} and Figure~\ref{fig:ahc025-budget} show best-so-far performance against billed model
spend on AHC015, AHC025, and AHC041, measured on the public evaluation cases.
\method{} finishes highest on all three, with a clear lead.

\FloatBarrier

\section{Search Strategies Written by the Agents}
\label{app:strategies}

\newcommand{\qask}[1]{\par\noindent\emph{#1}\par\smallskip}
\newenvironment{cypher}{\fontsize{8}{9.5}\selectfont\verbatim}{\endverbatim}

\method{} does not fix a search algorithm; the agents decide how to search by querying the
graph of past work. To see which strategies they actually use, we collected every Cypher query
the agents wrote across all runs (more than 2,500 queries). Below we show eight representative
queries (Boxes~\ref{box:q1}--\ref{box:q8}) and explain in plain terms what each one does. Queries
are copied from the logs; we changed only line breaks and replaced run keys that contain author
names with \texttt{anon/\ldots}.

As a reminder of the schema (Appendix~\ref{app:implementation}), an agent \texttt{RAN} sessions,
a session \texttt{PRODUCED} candidates, a candidate \texttt{USES\_IDEA} an idea, and a candidate is
\texttt{DERIVED\_FROM} its parents. Sessions store the lesson written at their end; candidates
store the program (\texttt{genome}), the score, every evaluator metric, and the agent's own
prediction of its score.

\textbf{Exploration and exploitation from one query.} The query the agents write most often
joins each idea to the candidates that used it and returns, per idea, how many times it was tried
and the best score it reached (Box~\ref{box:q1}). Sorted with the least-tried ideas first, it
answers an exploration question: which directions were proposed but barely tested, and did the
one attempt look promising? Sorted by best score instead, the same query answers the exploitation
question. This is the information a bandit rule such as UCB uses (visit counts and values), but
the agent chooses the trade-off itself rather than through a fixed formula.

\textbf{Leaving a crowded basin.} When one family of ideas holds the whole leaderboard, the agent
filters that family out by name and asks how good the best approach \emph{outside} it is
(Box~\ref{box:q2}). On Cloudcast, every top candidate was a Steiner-tree variant; the agent used
this query to decide whether a different family was worth a candidate.

\textbf{Choosing a parent from the lineage.} To decide what to build on, the agent follows the
\texttt{DERIVED\_FROM} edges backward from the best candidates (Box~\ref{box:q3}). On MMD-14-3,
every candidate above 0.99 came from the same parent, so the agent took that parent as its
starting point instead of the current best candidate.

\textbf{Reading code as data.} Because programs are stored in the graph, the agent can filter
candidates by the text of their code and extract values from it inside the query
(Box~\ref{box:q4}). On circle packing, one call returned the circle coordinates of every good
packing, without opening a single file.

\textbf{Advice that was never acted on.} Some queries are saved as named tools and reused in later
sessions. The most reused one in the Erd\H{o}s overlap run searches the lessons of all past
sessions for algorithm names (Remez, Frank--Wolfe, SQP, Newton, and others) and reports how many
sessions recommended each (Box~\ref{box:q5}). Compared with the candidates, this list shows
algorithm classes that several sessions had recommended but no candidate had tried. Neither the
lessons alone nor the leaderboard alone shows this; it takes a query over both.

\textbf{When was the last breakthrough?} Another query rebuilds the running record in submission
order and keeps only the candidates that broke it, together with their ideas (Box~\ref{box:q6}).
It shows at a glance how long ago the record last moved and which ideas produced jumps rather
than small refinements.

\textbf{Trading off two metrics.} On LLM-SQL, the score combines hit rate and runtime. One query
keeps only candidates that no other candidate beats on both, the Pareto frontier
(Box~\ref{box:q7}), which tells the agent whether there is more to gain in accuracy or in speed.

\textbf{Making failed experiments useful.} The last example comes from a channel-coding problem
that we ran independently of this paper; it is not one of our benchmarks. The query lists every
candidate that was
submitted knowingly \emph{below} the record, with how much score it cost and the hypothesis it was
meant to test (Box~\ref{box:q8}). On the leaderboard these probes look like failures; this query
shows them as measurements that later sessions can build on.

Other queries in the same spirit return the best candidate per idea (the MAP-Elites selection
rule) and the gain of every child over its parent along the lineage edges.

\colorlet{promptfield}{neagentbg}
\begin{promptbox}[searchprompt,label=box:q1]{Least-tried ideas first (LLM-SQL)}
\qask{Which ideas have been named but barely tested, and how good was their best attempt?}
\begin{cypher}
MATCH (i:Idea)
OPTIONAL MATCH (c:Candidate)-[:USES_IDEA]->(i)
WITH i, count(c) AS n_candidates,
     CASE WHEN count(c) > 0 THEN max(c.combined_score) ELSE null END
       AS best_score
RETURN i.name, i.description, n_candidates, best_score
ORDER BY n_candidates ASC, best_score DESC
LIMIT 20
\end{cypher}
\end{promptbox}

\begin{promptbox}[searchprompt,label=box:q2]{Best candidate outside the dominant family (Cloudcast)}
\qask{Excluding tree-based ideas, what is the best and worst result of each remaining idea?}
\begin{cypher}
MATCH (c:Candidate {eval_status: 'done'})-[:USES_IDEA]->(i:Idea)
WHERE NOT i.name CONTAINS 'steiner' AND NOT i.name CONTAINS 'mst'
  AND NOT i.name CONTAINS 'tree'
RETURN i.name, COUNT(c) AS count, MIN(c.objective) AS best,
       MAX(c.objective) AS worst
ORDER BY best LIMIT 20
\end{cypher}
\end{promptbox}

\begin{promptbox}[searchprompt,label=box:q3]{Parents of the top candidates (MMD-14-3)}
\qask{Which parents did the best candidates come from?}
\begin{cypher}
MATCH (c:Candidate)-[:DERIVED_FROM]->(p:Candidate)
WHERE c.fitness > 0.99
RETURN c.genome_hash, c.fitness, p.genome_hash, p.fitness
ORDER BY c.fitness DESC
\end{cypher}
\end{promptbox}

\begin{promptbox}[searchprompt,label=box:q4]{Code as data (circle packing)}
\qask{Among good programs that return a packing and do not time themselves, what are the circle
coordinates written in the code?}
\begin{cypher}
MATCH (c:Candidate)
WHERE c.fitness > 0.9 AND c.genome CONTAINS "return circles"
  AND NOT c.genome CONTAINS "import time"
WITH c, [l IN split(c.genome, "\n")
         WHERE l STARTS WITH " [0." OR l STARTS WITH " [-"] AS rows
RETURN c.genome_hash AS h, c.fitness AS fit, rows
\end{cypher}
\end{promptbox}

\begin{promptbox}[searchprompt,label=box:q5]{\texttt{untried\_algorithm\_classes} (Erd\H{o}s overlap; called 5 times)}
\qask{Which algorithm classes did past lessons recommend, and how many sessions recommended each?}
\begin{cypher}
MATCH (s:Session)
WHERE s.lesson_learned IS NOT NULL AND s.id STARTS WITH 'anon/opus46-e1'
WITH collect({id: s.id, lesson: s.lesson_learned}) AS sessions
WITH sessions,
     ['Remez', 'exchange', 'Frank-Wolfe', 'SQP', 'ADMM', 'trust-constr',
      'active-set', 'dual problem', 'dual formulation', 'Newton', 'Hessian',
      'CMA-ES.*N=200', 'CMA-ES.*N=500', 'mathematical structure',
      'symmetr'] AS patterns
UNWIND patterns AS pattern
WITH pattern, sessions,
     [s IN sessions WHERE s.lesson CONTAINS pattern | s.id]
       AS recommending_sessions
WHERE size(recommending_sessions) > 0
OPTIONAL MATCH (c:Candidate) WHERE c.fitness IS NOT NULL
WITH pattern, recommending_sessions,
     max(c.fitness) AS global_best_fitness,
     min(c.c5_bound) AS global_best_c5
RETURN pattern AS algorithm_keyword, recommending_sessions,
       size(recommending_sessions) AS times_recommended,
       global_best_c5 AS current_best_c5
ORDER BY times_recommended DESC
\end{cypher}
\end{promptbox}

\begin{promptbox}[searchprompt,label=box:q6]{\texttt{score\_progression} (LLM-SQL)}
\qask{Which candidates broke the record, in which session, and with which ideas?}
\begin{cypher}
MATCH (a:Agent {key: 'anon/opus46-a'})-[:RAN]->(s:Session)
      -[:PRODUCED]->(c:Candidate)
WHERE c.eval_status = 'done' AND c.combined_score IS NOT NULL
WITH c, s ORDER BY c.created_at ASC
WITH collect({score: c.combined_score, hr: c.average_hit_rate,
              runtime: c.average_runtime, iteration: s.iteration,
              created: c.created_at, hash: c.genome_hash})
       AS all_candidates
WITH all_candidates, [i IN range(0, size(all_candidates)-1) |
  CASE WHEN i = 0 OR all_candidates[i].score >
       reduce(best = 0.0, j IN range(0, i-1) |
              CASE WHEN all_candidates[j].score > best
                   THEN all_candidates[j].score ELSE best END)
  THEN all_candidates[i] ELSE null END] AS filtered
UNWIND [x IN filtered WHERE x IS NOT NULL] AS record
OPTIONAL MATCH (cand:Candidate {genome_hash: record.hash})
               -[:USES_IDEA]->(idea:Idea)
RETURN record.score AS combined_score, record.hr AS hit_rate,
       record.runtime AS avg_runtime, record.iteration AS session,
       record.created AS when, collect(idea.name) AS ideas
ORDER BY record.score ASC
\end{cypher}
\end{promptbox}

\begin{promptbox}[searchprompt,label=box:q7]{\texttt{hr\_vs\_runtime\_frontier} (LLM-SQL)}
\qask{Which candidates are not beaten by any other candidate on both hit rate and runtime?}
\begin{cypher}
MATCH (c:Candidate {eval_status: 'done'})
WITH c.average_hit_rate AS hr, c.average_runtime AS rt,
     c.combined_score AS cs, c.genome_hash AS gh
ORDER BY hr DESC
WITH collect({hr: hr, rt: rt, cs: cs, gh: gh}) AS all_candidates
UNWIND all_candidates AS cand
WITH cand, all_candidates
WHERE NONE(other IN all_candidates
           WHERE other.hr >= cand.hr AND other.rt <= cand.rt
             AND other.gh <> cand.gh)
RETURN cand.gh AS genome_hash, cand.hr AS hit_rate, cand.rt AS runtime,
       cand.cs AS combined_score
ORDER BY cand.hr DESC
\end{cypher}
\end{promptbox}

\begin{promptbox}[searchprompt,label=box:q8]{\texttt{measured\_negatives} (channel-coding task)}
\qask{Which candidates scored below the record at the time, by how much, and what did they test?}
\begin{cypher}
MATCH (a:Agent {key: "anon/opus5-j1"})-[:RAN]->(s:Session)
      -[:PRODUCED]->(c:Candidate)
WITH collect({it: s.iteration, f: c.fitness, g: c.gap_db,
              lp: c.linguistic_prediction}) AS all
UNWIND all AS x
WITH all, x, [y IN all WHERE y.it < x.it | y.f] AS earlier
WITH x, reduce(m = 0.0, v IN earlier | CASE WHEN v > m THEN v ELSE m END)
       AS record_before
WHERE record_before > 0.0 AND x.f < record_before - 0.0005
RETURN x.it AS iteration,
       round(x.f * 1000000) / 1000000 AS scored,
       round(record_before * 1000000) / 1000000 AS record_at_the_time,
       round((record_before - x.f) * 1000000) / 1000000 AS cost_in_score,
       x.g AS per_snr_gaps, x.lp AS what_it_measured
ORDER BY iteration, cost_in_score
\end{cypher}
\end{promptbox}

\textbf{What the queries show.} The agents do use the ingredients of classical search algorithms,
but not as a fixed rule. Counting how often each idea was tried and how well it did
(Box~\ref{box:q1}) is the input to a bandit rule; the best candidate per idea is MAP-Elites;
following lineage back to a strong parent (Box~\ref{box:q3}) is the selection step of tree
search; and the Pareto frontier (Box~\ref{box:q7}) is the selection rule of multi-objective
evolutionary algorithms. The agents pick among these as the run demands. Other queries have no
counterpart in a fixed algorithm, because they read the agents' own text and code: matching
lesson prose against the candidates (Box~\ref{box:q5}), extracting numbers from programs
(Box~\ref{box:q4}), and recovering the hypotheses behind deliberately weak candidates
(Box~\ref{box:q8}). These need a model that reads the whole history, which is what \method{}
gives the agents.

\section{Results Averaged over Repeated Runs}
\label{app:repeats}

Each entry in the main tables is a single run. To check that the gains do not come from the
randomness of a single run, we repeated a subset of the tasks: three runs of \method{} and CORAL
on the single-cell denoising tasks and on AHC006, AHC012, and AHC015
(Table~\ref{tab:repeats-ttt}), two runs of every method on the kernel builder, and three runs of
\method{} on transaction scheduling and Cloudcast (Table~\ref{tab:repeats-kernel}). The tables
report the mean over runs, with the individual runs in parentheses.

\method{} has the better mean on every task. On five of the nine tasks (pancreas, Tabula
lung, AHC015, the kernel builder, and Cloudcast), even its worst run is better than the best run
of every baseline, and on transaction scheduling its worst run ties CORAL. The improvement is
therefore not the result of a lucky run.

The spend--performance curves of Appendix~\ref{app:spend-curves} point the same way. On every
task shown there, the best-so-far curve of \method{} lies above those of the baselines at almost
every level of spend: at a given budget it reaches a higher score, and it reaches a given score
at lower spend. The second run of the kernel builder (Figure~\ref{fig:ahc025-budget}b) repeats
the pattern of the first (Figure~\ref{fig:kernel}). Run-to-run randomness therefore plays only a
small role compared with the gap between \method{} and the baselines.

\begin{table}[!ht]
\centering
\small
\caption{Repeated runs of \method{} and CORAL on the single-cell denoising tasks and three
AtCoder contests. Mean over three runs, individual runs in parentheses; better mean in bold.}
\label{tab:repeats-ttt}
\setlength{\tabcolsep}{5pt}
\resizebox{\linewidth}{!}{%
\begin{tabular}{lll}
\toprule
Task & \method{} & CORAL \\
\midrule
Pancreas (search split) $\downarrow$ & \textbf{0.1607} (0.1603, 0.1602, 0.1616) & 0.1685 (0.1676, 0.1724, 0.1654) \\
PBMC (held out) $\downarrow$ & \textbf{0.1495} (0.1472, 0.1473, 0.1539) & 0.1555 (0.1559, 0.1571, 0.1534) \\
Tabula lung (held out) $\downarrow$ & \textbf{0.1323} (0.1320, 0.1308, 0.1342) & 0.1379 (0.1392, 0.1386, 0.1360) \\
\midrule
AHC006 $\uparrow$ & \textbf{2,319,175} (2,324,852, 2,323,655, 2,309,017) & 2,282,891 (2,317,352, 2,297,709, 2,233,612) \\
AHC012 $\uparrow$ & \textbf{99,991,143} (99,997,290, 99,996,203, 99,979,937) & 99,936,491 (99,977,717, 99,847,640, 99,984,115) \\
AHC015 $\uparrow$ & \textbf{167,284,540} (168,660,525, 165,531,643, 167,661,453) & 156,515,263 (156,230,796, 155,943,176, 157,371,816) \\
\bottomrule
\end{tabular}}
\end{table}

\begin{table}[!ht]
\centering
\small
\caption{Repeated runs on the kernel builder (two runs of every method, cycle count at the
\$118 budget) and on transaction scheduling and Cloudcast (three runs of \method{}). Mean over
runs, individual runs in parentheses; best mean in bold. The baselines on transaction scheduling
and Cloudcast are the results reported by \citet{qu2026coral} (Table~\ref{tab:coral}).}
\label{tab:repeats-kernel}
\setlength{\tabcolsep}{4pt}
\resizebox{\linewidth}{!}{%
\begin{tabular}{lccccc}
\toprule
Task & \method{} & OpenEvolve & ShinkaEvolve & EvoX & CORAL \\
\midrule
Kernel builder (cycles, \$118) $\downarrow$ & \textbf{985} (960, 1,010) & 1,182 (1,174, 1,190) & 1,404 (1,156, 1,652) & 1,118.5 (1,098, 1,139) & 1,149.5 (1,148, 1,151) \\
Txn Sched.\ $\uparrow$ & \textbf{4622.7} (4651, 4566, 4651) & 3774 & 3802 & 3984 & 4566 \\
Cloudcast $\downarrow$ & \textbf{618.0} (618.0, 618.0, 618.0) & 627.2 & 627.8 & 623.5 & 618.4 \\
\bottomrule
\end{tabular}}
\end{table}
\FloatBarrier

\section{Implementation Details}
\label{app:implementation}

This appendix adds implementation details to the components of \S\ref{subsec:framework}.
Prompt templates and tool descriptions are in Appendix~\ref{app:prompts}, and the settings of our
experiments in Appendix~\ref{app:details}.

\subsection{Architecture and Run Lifecycle}

\paragraph{Processes and setup.}
The trusted side is the orchestrator, a Neo4j database whose data persists between runs, and the
MCP server, a host process. Only this trusted code holds the database password, and the database
and the server listen on loopback only.
Containers are started through one sandbox module that supports Docker, Podman,
Apptainer/Singularity, and Enroot.
Before any agent starts, the orchestrator starts the database and the MCP server and builds the
evaluation environment.
Each branch takes its starting iteration from the database, so a new invocation continues a branch
where it stopped.
A new branch's worktree (its workspace in \S\ref{subsec:framework}) is a fresh git repository, not connected to the main repository's history,
holding a copy of the problem's \texttt{solution.py} and empty \texttt{tools/} and \texttt{skills/}
folders.

\paragraph{Iterations.}
Each branch runs its sessions in sequence in its own loop, in parallel with the other branches;
iterations are grouped into blocks, and all branches wait at a barrier at the end of a block before
the meta pass.
To open a session, the orchestrator renders the instruction file from a template with the branch's
current guidance last, builds a short briefing from fixed queries (Appendix~\ref{app:prompts}), and
creates the \texttt{Session} node.
The coding-agent CLI then runs non-interactively (\texttt{claude -p} or \texttt{codex exec}) in the
branch's container.
The session ends when the CLI exits or its wall clock expires; a crash or timeout is recorded and
the run continues.
On close, the orchestrator stores the reasoning text from the transcript on the \texttt{Session}
node, writes the exposure edges (\S\ref{app:impl-schema}), and commits the worktree.

\paragraph{State across sessions.}
A branch keeps its worktree (with \texttt{solution.py} as the last session left it,
\texttt{tools/}, and \texttt{skills/}), the worktree's git history, its guidance, the tools the meta
agent wrote for it, and the shared graph.
Each session is a new CLI process with a re-rendered instruction file.
One exception: a session cut off because the model provider's usage window ran out is resumed
with its transcript once the window resets.

\subsection{Knowledge Graph Schema}
\label{app:impl-schema}

Tables~\ref{tab:schema-nodes} and~\ref{tab:schema-edges} list the node labels and relationships.
Every key has a uniqueness constraint.
A candidate's key is the SHA-256 of its program text, so a resubmitted program never becomes a
second node; its \texttt{fitness} is the evaluator's score, and each further metric the evaluator
reports is a property of its own.

\begin{table}[t]
\centering
\footnotesize
\caption{Node labels of the knowledge graph, with what each node holds.}
\label{tab:schema-nodes}
\begin{tabular}{@{}llp{0.6\linewidth}@{}}
\toprule
Label & Key & Holds \\
\midrule
\texttt{User}      & \texttt{name} & \raggedright The person who owns the agents. \tabularnewline
\texttt{Agent}     & \texttt{key} & \raggedright One branch: a persistent search-agent identity and the
                     path of its worktree. \tabularnewline
\texttt{Session}   & \texttt{id} & \raggedright One search-agent session: its iteration, model and
                     briefing, its reasoning text (\texttt{thinking\_text}), start and end times, the
                     fingerprint of its call token (\texttt{token\_hash}), its own lesson
                     (\texttt{lesson\_learned}), and the meta agent's judgment of it
                     (\texttt{meta\_lesson}). \tabularnewline
\texttt{Candidate} & \texttt{genome\_hash} & \raggedright One unique program (\texttt{genome}, the
                     full \texttt{solution.py}), its \texttt{fitness} and evaluation status, one
                     property per evaluator metric, and the agent's predictions, made before
                     scoring. \tabularnewline
\texttt{Idea}      & \texttt{name} & \raggedright A named idea, reused across candidates, with a
                     description. \tabularnewline
\texttt{Resource}  & \texttt{name} & \raggedright A file the problem provides, and where agents find
                     it. \tabularnewline
\texttt{Guidance}  & \texttt{id} & \raggedright One version of a branch's strategy, with a summary of
                     what changed; the first is a seed, and later ones are written by the meta
                     agent. \tabularnewline
\texttt{Tool}      & (\texttt{name}, \texttt{branch}) & \raggedright A read-only query the meta
                     agent saved for one branch: its description, Cypher code, and input
                     schema. \tabularnewline
\texttt{RunConfig} & \texttt{problem} & \raggedright The settings the current run was launched
                     with, which every process reads back. \tabularnewline
\bottomrule
\end{tabular}
\end{table}

\begin{table}[t]
\centering
\footnotesize
\caption{Knowledge-graph relationships and their writers. Endpoints are shown as
source $\rightarrow$ target; \emph{facts} are reconstructed from the session log, whereas
\emph{claims} are the counterpart declared by the search agent.}
\label{tab:schema-edges}
\setlength{\tabcolsep}{4pt}
\renewcommand{\arraystretch}{1.05}
\begin{tabular}{@{}ll@{}}
\toprule
Writer / provenance & Relationships \\
\midrule
orchestrator & \texttt{OWNS} (User$\rightarrow$Agent), \texttt{RAN} (Agent$\rightarrow$Session), \\
 & \texttt{UNDER\_GUIDANCE} (Session$\rightarrow$Guidance), seed \texttt{HAS\_GUIDANCE} (Agent$\rightarrow$Guidance) \\
orchestrator (\emph{facts}) & \texttt{RETRIEVED} \{\texttt{via}\} (Session$\rightarrow$Candidate),
  \texttt{READ} (Session$\rightarrow$Resource) \\
\addlinespace[1pt]
server & \texttt{PRODUCED} (Session$\rightarrow$Candidate), \texttt{USED\_TOOL} (Session$\rightarrow$Tool) \\
\addlinespace[1pt]
search agent & \texttt{USES\_IDEA} (Candidate$\rightarrow$Idea), \texttt{RELATED\_TO} \{\texttt{how}\} (Idea$\rightarrow$Idea) \\
search agent (\emph{claims}) & \texttt{DERIVED\_FROM} \{\texttt{kind}\} (Candidate$\rightarrow$Candidate), \\
 & \texttt{USED\_CONTEXT} (Candidate$\rightarrow$Resource) \\
\addlinespace[1pt]
meta agent & \texttt{HAS\_GUIDANCE} (Agent$\rightarrow$Guidance), \texttt{HAS\_TOOL} (Agent$\rightarrow$Tool), \\
 & \texttt{SUPERSEDES} (Guidance$\rightarrow$Guidance) \\
\bottomrule
\end{tabular}
\end{table}

\paragraph{Facts and claims.}
Edges written by the orchestrator or the server record what happened; the agent cannot write them.
Edges written by the search agent are its declarations, which the server checks only for existence.
Two declarations have a logged counterpart: a candidate's \texttt{DERIVED\_FROM} and
\texttt{USED\_CONTEXT} edges claim which candidates and files it came from, while the session's
\texttt{RETRIEVED} and \texttt{READ} edges, reconstructed by code from its transcript, record which
it actually fetched.

\subsection{Meta Pass}

The optional meta pass runs at the barrier between blocks; a single meta agent serves all branches, and if
the pass fails, the next block keeps the guidance already in place.
Its container mounts each branch's transcripts read-only and gets neither the evaluator, the
database password, any \texttt{solution.py}, nor any git history.
Besides the skill files of \S\ref{subsec:framework}, it writes through three tools of its own:
\texttt{write\_meta\_lesson} stores its judgment of one session on that \texttt{Session} node;
\texttt{add\_tool} stores a read-only saved query for one branch, which the branch's later sessions
are offered as an MCP tool; and \texttt{write\_strategy} stores a
branch's new strategy as a \texttt{Guidance} node that supersedes the previous one and is rendered
into the branch's next instruction file.

\subsection{Execution Safeguards}
\label{app:safeguards}

\begin{itemize}
    \item \textbf{Evaluator isolation.} The evaluator is never mounted into an agent's container,
    and a worktree's git history cannot contain it. The server reads \texttt{solution.py} from the
    agent's worktree itself and evaluates a copy in a separate container, under limits set per
    problem and without network access (except on the Enroot and \texttt{none} backends, which
    cannot disable it). Hidden
    evaluation data is fetched before any candidate exists, into a cache no agent can see.

    \item \textbf{Agent containers.} A search agent's container holds its own worktree and the
    problem's public files, and never the database password. The CLI's web tools are removed by default,
    and the prompt forbids looking up results online (Appendix~\ref{app:prompts}); the container keeps the network access the CLI
    needs to reach its model API.

    \item \textbf{Read-only queries, typed writes.} \texttt{read\_graph} runs in a Neo4j read
    transaction, so the database itself rejects any write; all writes go through the fixed tools.

    \item \textbf{Attribution.} The server attributes every call to a session itself, never
    trusting what an agent says about itself, and offers each caller only its own tools.
\end{itemize}

\section{Agent Prompts and Tools}
\label{app:prompts}

\definecolor{promptfield}{gray}{0.88}
\newcommand{\phead}[1]{\par\smallskip\noindent{\bfseries #1}\par\vspace{1pt}}
\newcommand{\pstep}[2]{\par\noindent\hangindent=1.65em\hangafter=1%
  \makebox[1.65em][l]{\bfseries #1.}#2}
\newcommand{\tfield}[1]{{\setlength{\fboxsep}{1.2pt}\colorbox{promptfield}{%
  \strut\texttt{\char`\{\char`\{#1\char`\}\char`\}}}}}
\newcommand{\rfield}[1]{{\setlength{\fboxsep}{1.2pt}\colorbox{promptfield}{%
  \strut\itshape\ensuremath{\langle}\,#1\,\ensuremath{\rangle}}}}
\newcommand{\searchcaller}{\textcolor{neagent}{\bfseries search}}
\newcommand{\metacaller}{\textcolor{nemeta}{\bfseries meta}}
\newcommand{\bothcaller}{\textcolor{neruntime}{\bfseries both}}

The search and meta agents receive different instructions and tool sets. The boxes below preserve
the wording of the consequential passages while condensing long lists.

\subsection{Search-Agent Instructions}
\label{app:prompts-search}

At the start of each search session, the orchestrator creates \texttt{CLAUDE.md} or
\texttt{AGENTS.md} by substituting \tfield{PROBLEM}, \tfield{SCHEMA},
\tfield{FOLDER\_LISTING}, and \tfield{STRATEGY} in one template. Box~\ref{box:search}
selects the parts that define the agent's objective, interface, persistent state, and search loop.

\colorlet{promptfield}{neagentbg}
\begin{promptbox}[searchprompt,label=box:search]{Selected search-agent instructions (abridged)}
\phead{AgentDiscover}
You are working on an evolutionary search over \texttt{solution.py}, building on work that is
already done. A shared database holds every candidate submitted so far, the ideas behind them and
the lessons written about them; your working directory holds \texttt{tools/} and \texttt{skills/}
left by earlier work. Explore the database and those folders first, then either build on what you
find or pursue your own idea.

\phead{Problem}
\tfield{PROBLEM}

\phead{What you are judged by}
Your briefing names a \textbf{target}. Beating it, preferably by a clear margin, is what this run is
for. Your job is to push the best number in the database past it, as far past as you can get within
this session's cap.

The target is also your yardstick. After each candidate you see your score and the distance still
to go: judge from that whether your current idea can realistically reach the target, or whether it
is levelling off short of it. If it is levelling off, change idea rather than refine it further.

\phead{Your tools}
\texttt{read\_graph(cypher)}---run any read-only Cypher query and get rows as JSON.\par
\texttt{add\_idea(name, description, related\_to)}---record the idea a candidate will use, or
reuse an existing idea by exact name.\par
\texttt{submit\_candidate(idea, linguistic\_prediction, fitness\_prediction, parents, resources)}
---submit the current \texttt{solution.py}; the server evaluates it in a locked sandbox and returns
the score.\par
\texttt{write\_lesson(text)}---after your last candidate, write one lesson for the whole session.

Your whole working directory persists and is committed after every session. Put reusable code in
\texttt{tools/}; \texttt{skills/} contains notes, written from the evidence so far, on how the
search on this problem has gone.

\phead{The graph you can read}
This is a graph database, not a table: \texttt{read\_graph} takes any read-only Cypher, so follow
relationships across several hops rather than only listing nodes.

\tfield{SCHEMA}

\phead{The session loop}
Your prompt names the maximum number of candidates for this session. It is a cap, not a quota.
Push the fitness with every candidate. If your improvements look like they are levelling off short
of the target, switch to a different idea rather than tweak the same one further.

\pstep{1}{\textbf{Explore}---briefing first, then \texttt{read\_graph}. Before your first candidate,
read the most recent lessons; explore beyond that as much as is useful.}
\pstep{2}{\textbf{An idea} (mandatory)---\texttt{add\_idea} a new one, or pick an existing idea's
exact name to reuse.}
\pstep{3}{\textbf{A candidate} (mandatory)---edit only the marked evolve block of
\texttt{solution.py}; test locally; decide its parents; write both predictions
before submitting; then call \texttt{submit\_candidate} and read the score it returns.}
\pstep{4}{Loop back, building on what the scores taught you.}

When you stop---at the cap, or earlier because different ideas all failed---call
\texttt{write\_lesson} with one lesson covering the whole session. Then stop; the orchestrator
takes over from there.

\phead{Honesty rules}
Your declared parents, idea, and resources are recorded as your claims about how the candidate was
made. What you actually retrieved and read is recorded separately, by the platform, from your
session log.

\phead{Persistent artifacts and current strategy}
\tfield{FOLDER\_LISTING}\par
\tfield{STRATEGY}
\end{promptbox}

\subsection{Meta-Agent Instructions}
\label{app:prompts-meta}

After each block, the orchestrator renders the meta template by substituting the problem and graph
schema, the block and session sizes, branch paths, and a block summary. The meta agent sees the
search traces and outcomes but not the evaluator. Box~\ref{box:meta} selects the
instructions that define its evidence, outputs, and boundary with the search agent.

\colorlet{promptfield}{nemetabg}
\begin{promptbox}[metaprompt,label=box:meta]{Selected meta-agent instructions (abridged)}
\phead{Role}
You are the meta-learning step, not one of the search branches. You run after all search agents
have paused for this block. Search agents are short-lived: each is spawned fresh with no memory,
proposes a few candidate \texttt{solution.py} programs, and stops. Your job is to find
\textbf{which ways of thinking paid off}, and to make them cheap for the next fresh agents.

Whatever you write for a branch stays in force for \tfield{META\_EVERY} sessions, each proposing
\tfield{PROPOSALS} candidates. That horizon tells you how long your claims must keep being true;
it is not a budget for you to allocate.

\phead{What you can reach}
The shared history graph through \texttt{read\_graph}; each branch's raw session logs (read-only);
and each branch's \texttt{skills/} directory (writable). You do not have any genome,
\texttt{solution.py}, the evaluator, or database credentials. You shape \emph{how} branches search,
never the genomes directly.

Branches: \tfield{BRANCHES}\par
Briefing for this block: \tfield{BRIEFING}

\phead{What to read}
The useful material is always \textbf{pairs}---thinking joined with outcome, because thinking alone
is stories and fitness alone is numbers. Compare session reasoning with candidate fitness,
predictions with results, idea and lineage reuse, tool use, and performance under earlier guidance.

\phead{What you write---per branch}
\pstep{1}{\texttt{write\_meta\_lesson(session\_id, text)}---for each session you reviewed: your
judgment of how well it searched, independent of the score it got.}
\pstep{2}{\texttt{add\_tool(...)}---a \textbf{saved query}, and only that: a read-only Cypher
template exposed to later sessions as a named tool.}
\pstep{3}{\textbf{Skills}---short Markdown files of methodological guidance from behavioural
evidence: how the branch tends to fail, what exploration paid for, and what need not be re-derived.}
\pstep{4}{\texttt{write\_strategy(...)}---what ways of thinking pay off for this branch and what
ways do not, written to the next fresh agent.}

\phead{The line between your job and theirs}
\textbf{Deciding what to try next is the search agent's job. Deciding how to think about the
problem is yours.} Do not write candidate allocations, ranked lists of approaches, or
implementation into a strategy. A test before you write: \emph{would this still make sense to an
agent that finds something you did not anticipate?}
\end{promptbox}

\subsection{Session Briefing and Feedback}
\label{app:prompts-reply}

The orchestrator's command-line prompt for the search agent adds a candidate cap, the internet policy, and a briefing built by
fixed database queries. With web access off (the default), the internet rule reads ``Never reach
out to the internet in any way (web search, curl, wget, git clone, or otherwise) to look up other
results for this problem.''; otherwise ``You may search the internet if you want.'' Each submission
returns a code-generated steering message from the MCP server. Box~\ref{box:reply} shows their stable forms and the
cost note; angle-bracketed fields are runtime values, and other optional variants are omitted.

\colorlet{promptfield}{neruntimebg}
\begin{promptbox}[runtimeprompt,label=box:reply]{Search-session prompt and server feedback (abridged)}
\phead{Invocation prompt}
Submit up to \rfield{proposals} candidates with the \texttt{submit\_candidate} tool.

Do not reply with a plan or a summary of what you intend to do: a reply that calls no tool ends
your session immediately, and a session that submits nothing is wasted. Explore as much as you
need, then edit \texttt{solution.py} and submit.

\rfield{internet rule}

\phead{Briefing}
Best fitness in your branch: \rfield{value}\par
Best in the database: \rfield{value}\par
Best \rfield{objective name} in the database: \rfield{value}
\quad(target to beat clearly: \rfield{target})\par
The database holds \rfield{n} candidates and \rfield{m} ideas.\par
\rfield{k} search agents are running in parallel.\par
This whole session is capped at \rfield{wall} s (approximately \rfield{hours} h)---budget your
candidates against it; a session cut off at the cap loses its lesson.

\phead{Reply below the cap}
\rfield{objective name} = \rfield{value} (you predicted \rfield{prediction}) (target
\rfield{target}, \rfield{gap}\% away). fitness \rfield{fitness}. That was candidate \rfield{n} of at most \rfield{cap}. Earlier this session:
\rfield{earlier values with predictions}. Best this session: \rfield{value}. Best in the database:
\rfield{value}.

You can read the graph again, and use or add to your persistent \texttt{tools/} folder. Work on
your next candidate: refine what shows promise, or if these numbers say you are stuck, switch to a
different idea. Stop early only if you have tried different ideas and none worked---then write the
lesson. About \rfield{hours} h of this session's wall clock remain (\rfield{seconds} s).
\rfield{cost note}

\phead{Cost note (from the fifth candidate, when a fresh session would be cheaper)}
The token cost of continuing this session will probably be higher than the cost of starting a new
one (this session's context has grown to \rfield{L} tokens; a fresh session would start from
\rfield{c + i}). Continue only if you have a great idea that is very likely to give a good result;
otherwise write the lesson and terminate, and you will get another chance to try in a new session.

\phead{Reply at the cap}
That was candidate \rfield{n} of at most \rfield{cap}---the cap. Now write the lesson for this
session with \texttt{write\_lesson}---include the work that never became a candidate, since only
submitted ones are in the graph---then stop.
\end{promptbox}

\subsection{Tools}
\label{app:prompts-tools}

Table~\ref{tab:tools} lists the MCP interface. The ``saved query'' in the meta prompt is a named,
branch-specific, read-only Cypher query registered with \texttt{add\_tool}; we call it a
\emph{meta-written query tool} below. Later search sessions see its name, description, and argument
schema, but not its Cypher implementation.

\begin{table}[!ht]
\centering
\footnotesize
\caption{MCP tools available to search and meta agents.}
\label{tab:tools}
\setlength{\tabcolsep}{4pt}
\begin{tabular}{@{}llp{0.29\linewidth}p{0.36\linewidth}@{}}
\toprule
Tool & Caller & Arguments & Effect \\
\midrule
\texttt{read\_graph} & \bothcaller & \raggedright \texttt{cypher} & \raggedright Runs a read-only
  Cypher query and returns rows as JSON. \tabularnewline
\texttt{add\_idea} & \searchcaller & \raggedright \texttt{name}, \texttt{description},
  \texttt{related\_to} & \raggedright Creates or reuses an idea and records its relations to
  existing ideas. \tabularnewline
\texttt{submit\_candidate} & \searchcaller & \raggedright \texttt{idea}, predictions,
  \texttt{parents}, \texttt{resources} & \raggedright Evaluates the current \texttt{solution.py},
  records its provenance and result, and returns steering feedback. \tabularnewline
\texttt{write\_lesson} & \searchcaller & \raggedright \texttt{text} & \raggedright Stores the
  session's final lesson. \tabularnewline
\addlinespace
\texttt{write\_meta\_lesson} & \metacaller & \raggedright \texttt{session\_id}, \texttt{text} &
  \raggedright Stores the meta agent's assessment of one session's search process. \tabularnewline
\texttt{add\_tool} & \metacaller & \raggedright \texttt{branch}, \texttt{name},
  \texttt{description}, \texttt{code}, \texttt{params} & \raggedright Registers a read-only
  Cypher query as a tool for one branch. \tabularnewline
\texttt{write\_strategy} & \metacaller & \raggedright \texttt{branch}, \texttt{strategy},
  \texttt{diff\_summary} & \raggedright Replaces a branch's guidance for later sessions. \tabularnewline
\addlinespace
meta-written query tool & \searchcaller & \raggedright arguments declared by its schema &
  \raggedright Runs the branch-specific query registered by the meta agent. \tabularnewline
\bottomrule
\end{tabular}
\end{table}

\section{Evaluator Failure Modes and Corrections}
\label{app:evaluators}

The evaluation sandbox (Appendix~\ref{app:safeguards}) keeps the evaluator and its private
files outside the agent's workspace.  The submitted program must nevertheless run somewhere,
and the evaluator must determine whether its output is a valid solution.  While porting the
benchmark evaluators, we repeatedly encountered the following failure modes.  We report these
recurring patterns rather than attempt an exhaustive audit of every evaluator or upstream version.

\paragraph{Incomplete output validation.}
\emph{Affected tasks: Signal Processing, LLM-SQL.}
An evaluator may compute a favorable score for an output that does not satisfy the constraints
of the stated task if it checks only that the output can be parsed.  On Signal Processing, the
required output length was never enforced: a candidate that returned only two samples, with no
slope reversals and near-perfect correlation, scored $0.9259$ on the evaluator,
although it is not a solution to the task at all.  Even with the length enforced, the metric
rewards smoothness and penalizes the hidden clean signal for its own oscillations, so an
oversmoothed output can score above the exact ground truth, which scores $0.727$.  We leave
the metric itself unchanged, so that our scores stay comparable with the evaluator used in
prior work.  On LLM-SQL,
the returned table was never checked to be a rearrangement of the input rows: a candidate that
discarded the input and returned one long constant row $n$ times made every row after the first
a full prefix-cache hit and scored $0.994$, against $0.737$ for the best honest reordering we
found.  We therefore check the output's shape, contents, completeness, and resource constraints
against authoritative task data before scoring it.

\paragraph{Candidate-controlled scoring.}
\emph{Affected tasks: Signal Processing, 3rd-Autocorr., PRISM, and LLM-SQL result loading.}
If candidate code runs in the process that computes the score, it can rebind the metric
functions, inspect scorer state, or simply print a result in the evaluator's name.  On Signal
Processing, a candidate that replaced \texttt{pearsonr} with a function returning $10^{9}$ was
scored $2\times10^{8}$, and one that printed \texttt{\{"score": 1e9\}} and exited had that value
reported at face value.  Loading a candidate-controlled serialized object into the scoring
process creates the same boundary failure: on LLM-SQL the returned DataFrame was exchanged via
\texttt{pickle}, whose deserialization executes code the candidate chooses.  We run candidate
code separately, exchange only the proposed solution, restrict deserialization to the expected
data types, and compute the score in a process the candidate did not enter.

\paragraph{Information leakage across the boundary.}
\emph{Affected task: EPLB.}
EPLB scores a placement on the \emph{next} recorded workload, so the workload data must stay
hidden: a candidate that sees the future no longer has to predict it.  The agent nevertheless
found two ways to obtain it.  First, although the agent's own sandbox blocks the data, the
candidate program it submits ran where the file was readable, and candidates that read it
raised the balance score from $0.25$ to $0.44$; a keyword blocklist did not stop this, since a
file path can be built indirectly.  Second, whatever a candidate prints was returned to the
agent as feedback, and the same workloads are replayed in the same order on every evaluation.
Candidates printed their inputs, and later candidates hardcoded the answers, lifting the score
from a best honest $0.153$ to $0.201$ in our strongest run.  Candidate processes therefore
receive only public task inputs (the private data is replaced by an empty directory), and their
stdout and exception messages are withheld from search feedback.

\section{Experimental Details}
\label{app:details}

This appendix records the model and run configuration for each evaluation setting in
\S\ref{sec:experiments}.  All \method{} runs use the Apptainer/Singularity backend of the sandbox described in
Appendix~\ref{app:implementation}; only settings that vary by experiment are listed below.

\subsection{Models, Access, and System Budget}

Table~\ref{tab:details-models} reports the model, access policy, and system-level budget.  The
reasoning-effort setting is shared by the search and meta agents; Table~\ref{tab:details-meta}
reports the meta model and schedule.

\begin{table}[!ht]
\centering
\small
\setlength{\tabcolsep}{5pt}
\caption{Model, access, and system-level budget ceiling by evaluation setting.  ``On'' for AC1 and AC2 applies to both the CORAL and \method{} runs.}
\label{tab:details-models}
\begin{tabular}{llccl}
\toprule
Setting & Search model & Effort & Web & Budget ceiling \\
\midrule
Systems + math & Opus~4.6 & default & off & 100 candidates \\
AtCoder (5 contests) & Opus~5 & default & off & matched model spend \\
Multi-domain (AHC039, AHC058) & Opus~5 & default & off & matched model spend \\
Multi-domain (AC1, AC2) & Opus~5 & default & on & matched model spend \\
Multi-domain (TriMul, MLA-Decode, denoising) & GPT-6 Sol & default & off & matched model spend \\
Kernel optimization & Opus~5 & low & off & matched model spend \\
Parallel-agent ablation & Haiku~4.5 & default & off & matched wall time / model spend \\
\bottomrule
\end{tabular}
\end{table}

The systems and mathematical baselines are published results compiled by
\citet{qu2026coral}; all methods use Claude Opus~4.6.  For the five AtCoder contests, we run
OpenEvolve, EvoX, CORAL, and \method{} with Claude Opus~5 and score their programs on the
official system tests.  For the multi-domain tasks,
TTT-Discover retains its published training and compute setting, while CORAL and \method{} run
under the same model-spend cap: with GPT-6 Sol on TriMul, MLA-Decode, and denoising, and with
Claude Opus~5 on AHC039, AHC058, AC1, and AC2.  All five kernel methods use Claude
Opus~5 at low reasoning effort and the same evaluator. The graph-database ablation
(Table~\ref{tab:db-ablation}) replays eight sessions from the transaction scheduling and
Erd\H{o}s overlap runs with Claude Opus~4.6.

For baseline runs we conduct ourselves, OpenEvolve, ShinkaEvolve, and EvoX run through
SkyDiscover~\citep{liu2026skydiscover} with their default configurations, and CORAL uses its
default configuration.  We change only the model and budget settings required by the comparison
protocol; for settings without web access, we also disable web access in CORAL's sandbox
configuration.

Candidate budgets count submitted programs, including those whose evaluation fails or times out;
only valid candidates can be reported.  Model spend is computed consistently for every method (Appendix~\ref{detail_spend}).
It is not a raw token count, and includes meta-agent calls when enabled.  Kernel results use the
best-so-far candidate at common spend checkpoints; wall-clock time is not used as a proxy for
model usage. Evaluator corrections are described
in Appendix~\ref{app:evaluators}.

\subsection{Search Schedule and Stopping Rules}

Table~\ref{tab:details-search} reports the maximum scheduled search work.  Every setting runs a
single search branch, except the parallel-agent ablation, which compares one branch with four.
Each iteration runs one search session on every active branch.  Runs continue until the shared
system-level budget in Table~\ref{tab:details-models} is exhausted.  The maximum-iterations
setting in Table~\ref{tab:details-search} is a non-binding safety ceiling, not the comparison
budget.

A session ends when the agent finds that it has stalled, or that carrying its context costs more
than starting afresh.  Each session also has a candidate cap and a time cap; these are upper limits
that control context length, not the comparison budget, and reaching the candidate cap
starts another session when one remains.  The launcher may raise the configured search timeout
when it cannot accommodate the candidate cap and evaluator's per-candidate limit.  Candidate
evaluation has a separate task-specific timeout.

Each main-table entry is the best valid candidate from one run.  The parallel-agent ablation uses five repeats per condition and reports means and standard
errors; the meta-agent ablation forks each of five runs three times (15 forks), and each fork is
compared with the unguided run it was forked from (\S\ref{sec:additional}); its search agent is Haiku~4.5 and its meta agent is Opus~4.6.

\begin{table}[!ht]
\centering
\small
\setlength{\tabcolsep}{5pt}
\caption{Search schedule and wall-clock ceiling by evaluation setting.}
\label{tab:details-search}
\begin{tabular}{lccc}
\toprule
Setting & Cand.\ / session & Max iter. & Search timeout (s) \\
\midrule
Systems + math & 25 & 5 & 28,800 \\
AtCoder (5 contests) & 20 & 10 & 36,000 \\
Multi-domain (Opus~5 tasks) & 20 & 5 & 57,600 \\
Multi-domain (GPT-6 Sol tasks) & 5 & 10 & 28,800 \\
Kernel optimization & 10 & 10 & 28,800 \\
Parallel-agent ablation & 8 & 10 & 14,400 \\
Meta-agent ablation & 10 & 5 & 28,800 \\
\bottomrule
\end{tabular}
\end{table}
\FloatBarrier

\subsection{Meta-Agent Schedule}

Table~\ref{tab:details-meta} reports the meta model, schedule, and timeout. One meta pass runs after each completed search iteration.

\begin{table}[!ht]
\centering
\small
\setlength{\tabcolsep}{5pt}
\caption{Meta-agent configuration by evaluation setting.}
\label{tab:details-meta}
\begin{tabular}{llc}
\toprule
Setting & Meta model & Meta timeout (s) \\
\midrule
Systems + math & Opus~4.6 & 7,200 \\
AtCoder (5 contests) & Opus~5 & 3,600 \\
Multi-domain (Opus~5 tasks) & Opus~5 & 5,400 \\
Multi-domain (GPT-6 Sol tasks) & GPT-6 Sol & 7,200 \\
Kernel optimization & Opus~5 & 7,200 \\
Parallel-agent ablation & Haiku~4.5 & 3,600 \\
\bottomrule
\end{tabular}
\end{table}
\FloatBarrier

The meta timeout is an independent ceiling: a pass can finish earlier, and a timeout skips that
pass without terminating search.  No pass is launched after the final iteration because no later
session could use its guidance.

\subsection{Model-spend accounting}
\label{detail_spend}
We estimate model spend from the token usage recorded in
the coding-agent transcripts (Claude Code or Codex) for each task. For every assistant
message, we price input, output, cache-read, and cache-write
tokens at the published rates for the model used, accounting
separately for 5-minute and 1-hour cache writes on Claude models. We sum
message costs across all search and meta agents in a run to obtain cumulative
model spend. These values are estimates at list prices, not
invoice charges, because the runs used subscriptions.

\section{AtCoder Submissions and Scoring Details}
\label{app:ahc-budgets}

\paragraph{Official submissions.}
The \method{} programs evaluated on the five contests in Table~\ref{tab:ahc} are
available for \href{https://atcoder.jp/contests/ahc006/submissions/80755704}{AHC006},
\href{https://atcoder.jp/contests/ahc012/submissions/79333505}{AHC012},
\href{https://atcoder.jp/contests/ahc015/submissions/79333575}{AHC015},
\href{https://atcoder.jp/contests/ahc025/submissions/79333762}{AHC025}, and
\href{https://atcoder.jp/contests/ahc041/submissions/79473791}{AHC041}. The programs
used for the multi-domain tasks in Table~\ref{tab:ttt} are available
for \href{https://atcoder.jp/contests/ahc039/submissions/79686776}{AHC039} and
\href{https://atcoder.jp/contests/ahc058/submissions/79686783}{AHC058}.
The AC1 construction is submitted to the
\href{https://einsteinarena.com/problems/first-autocorrelation-inequality}{Einstein Arena leaderboard},
where it ranked first with a bound of $1.50274353$ on October~3, 2026.

\subsection{AHC025 Scoring}
\label{app:ahc025-scoring}

AHC025 is scored differently from the other contests in Table~\ref{tab:ahc} and Table~\ref{tab:ttt}. Each test case has a fixed
\emph{absolute score} (lower is better), but the standings rank programs within each case
and sum rank-based scores, so a program's \emph{relative score} changes as others submit.
Table~\ref{tab:ahc} therefore reports the fixed sum of absolute scores, which the linked submission page also shows, and ranks from a single
extended-standings snapshot.\footnote{Official AHC025 scoring rules:
\url{https://atcoder.jp/contests/ahc025/tasks/ahc025_a?lang=en}.}

Table~\ref{tab:ahc025-absolute} reports each method's fixed absolute score alongside
its relative score, including ShinkaEvolve, which we ran on AHC025 only; the contest
rank counts only original contest entrants whose
relative score exceeds the submission's. \method{} attains the lowest absolute score (19,173,103, against 25,923,844 for the strongest baseline, CORAL), and the ordering of methods is the same under both scores.

\begin{center}
\begin{minipage}{0.94\linewidth}
\refstepcounter{table}
\label{tab:ahc025-absolute}
\small
\noindent Table~\thetable: AHC025 submitted-program results. Absolute score is
fixed and lower is better. Relative score and ranks are from the same mutable
extended-standings snapshot as the ranks in Table~\ref{tab:ahc} and are higher-is-better.
\par\smallskip
\centering
\setlength{\tabcolsep}{4pt}
\begin{tabular}{lrrrr}
\toprule
Method & Absolute $\downarrow$ & Relative $\uparrow$ & Contest & Extended \\
\midrule
OpenEvolve   & 95,926,386 & 92,596,442,685 & 37th & 51st \\
EvoX         & 39,086,466 & 94,461,264,820 & 21st & 30th \\
ShinkaEvolve & 31,925,711 & 95,932,806,317 & 12th & 18th \\
CORAL        & 25,923,844 & 98,498,020,580 & 2nd  & 6th  \\
\method{}    & \textbf{19,173,103} & \textbf{98,762,866,188} & \textbf{1st} & \textbf{3rd} \\
\bottomrule
\end{tabular}
\end{minipage}
\end{center}

\makeatletter
\setlength{\@fptop}{0pt}
\makeatother

\section{Steering Messages vs.\ CORAL's Heartbeat}
\label{app:steering}

Both keep a long-running agent on course, but CORAL's heartbeat wastes about 10\% of its LLM
calls, while our steering messages waste none. We measured this in our CORAL runs on all tasks of
Tables~\ref{tab:ttt} and~\ref{tab:ahc}.

\textbf{CORAL.} After submitting a candidate, the agent waits for its score, checking every
0.2\,s, and resumes work as soon as the score arrives. CORAL's manager checks for new scores only
every 5\,s. Once it sees the score, it interrupts the agent (SIGINT) and restarts the same session
with a steering prompt. In the gap between the two, the agent has usually already sent its next
LLM call. The interrupt aborts this call while the provider is processing it: the tokens are
billed, but the reply never reaches the restarted agent.

\textbf{\method{}.} The steering message is simply the output of the MCP tool the agent calls to
submit a candidate (Appendix~\ref{app:prompts-reply}). The agent is never interrupted, and no call
is lost.

\end{document}